\documentclass[twocolumn]{svjour3}

\usepackage{graphicx}%
\usepackage{multirow}%
\usepackage{amsmath,amssymb,amsfonts}%
\usepackage{mathrsfs}%
\usepackage{soul}%
\usepackage[title]{appendix}%
\usepackage{xcolor}%
\usepackage{textcomp}%
\usepackage{manyfoot}%
\usepackage{booktabs}%
\usepackage{algorithm}%
\usepackage{algorithmicx}%
\usepackage{algpseudocode}%
\usepackage{listings}%
\usepackage{makecell}
\usepackage{adjustbox}
\usepackage{stfloats}
\usepackage[square,numbers]{natbib}
\usepackage{hyperref}

\usepackage{xspace}

\newcommand{\eg}{e.\,g.,\xspace}
\newcommand{\ie}{i.\,e.,\xspace}
\newcommand{\cf}{cf.\xspace}

\begin{document}

\title{Comprehensive Evaluation of Prototype Neural Networks}

\author{Philipp Schlinge \and Steffen Meinert \and Martin Atzmueller}
\institute{Philipp Schlinge \at
              Semantic Information Systems Group, Osnabrück University, Germany \\
              \email{philipp.schlinge@uos.de}
           \and
           Steffen Meinert \at
              Semantic Information Systems Group, Osnabrück University, Germany \\
              \email{steffen.meinert@uos.de}
           \and
           Martin Atzmueller \at
              Semantic Information Systems Group, Osnabrück University, and 
              Research Department Cooperative and Autonomous Systems, German Research Center for Artificial Intelligence (DFKI), Germany\\
              \email{martin.atzmueller@uos.de}
}
\date{}
\maketitle

\begin{abstract}
Prototype models are an important method for explainable artificial intelligence (XAI) and interpretable machine learning. In this paper, we perform an in-depth analysis of a set of prominent prototype models including \textit{ProtoPNet}, \textit{ProtoPool} and \textit{PIPNet}. For their assessment, we apply a comprehensive set of metrics. In addition to applying standard metrics from literature, we propose several new metrics to further complement the analysis of model interpretability. In our experimentation, we apply the set of prototype models on a diverse set of datasets including fine-grained classification, Non-IID settings and multi-label classification to further contrast the performance. Furthermore, we also provide our source code as an open-source library (\url{https://github.com/uos-sis/quanproto}), which facilitates simple application of the metrics itself, as well as extensibility -- providing the option for easily adding new metrics and models.
\end{abstract}


\section{Introduction}\label{sec1}

While there have been enormous advances in deep learning~\cite{wang2020recent,pouyanfar2018survey}, the challenge of developing truly interpretable models with accuracy comparable to their black-box counterparts remains unresolved~\cite{rudin2022interpretablechallenges,du2019techniques,zhang2021survey}. This is a prominent and active area of research connecting both interpretability and explainability of the respective models~\cite{AFKS:24}. In particular, in the domain of image-related tasks like classification, non-interpretable models typically outperform most other models or techniques. However, in high-risk domains often encountered in industry or the medical sectors, explainability and interpretability are key requirements~\cite{rudin2019stop,stiglic2020interpretability}, also enabling computational sensemaking~\cite{atzmueller2017declarative}.

Different post-hoc techniques have been developed in the past, such as SHAP and LIME~\cite{SHAP,ribeiro2016should}. These explain the decision of a model by calculating the importance of individual input features and can be applied to a large family of models. In the domain of image classification, saliency maps are a particular prominent subgroup of these model-agnostic methods, \eg Integrated Gradients, Grad-CAM and other variants explain the attribution of individual or groups of pixels for a given prediction~\cite{integrated-gradient,Grad-CAM}. Although, such model-agnostic techniques are widely used and can provide considerable insight into trained models, they are in themselves only post-hoc methods with several drawbacks, \eg~\cite{adebayo2018sanity} compared to a completely intrinsically interpretable model.

One possible solution to overcome some of these problems, has been the development of deep part-based prototype models, firstly presented in~\cite{Li_Liu_Chen_Rudin_2018} and~\cite{chen2019thislookslikethat}. A key advantage is their intrinsic interpretability. Part-based prototype models focus on learning a meaningful embedding space using an often pre-trained black-box network as feature extractor. A key distinction from standard black-box models is the aim to extract features that concentrate on specific sections of the image. This division of the image in the embedding space allows the network to learn prototypes that represent only particular parts of the object. The classification layer then utilizes the presence of a prototype in the image to calculate the prediction. To provide an explanation for a prediction, the prototypes that have the highest evidence are projected back to the input space through a saliency method.

However, users have different experiences with model explanations. Confirmation bias and subjective understanding make it hard to guess the prediction correctly \cite{kim2022hive} \cite{nguyen2021effectiveness}. This shows that these models are not yet reliable enough to use for AI-assisted decision-making. This encouraged us to survey existing assessments of explainability using a well-founded categorization.

Our research is based on and extends the work of Nauta et al.~\cite{nauta2023co12}. With an implementation of a substantial portion of the evaluation framework presented in~\cite{nauta2023co12}, we facilitate a thorough assessment of part-based prototype models. Our contributions are summarized as follows:

\begin{enumerate}
\item We expand the set of metrics referenced in~\cite{nauta2023co12} by proposing several novel metrics designed to enhance the interpretability assessment of part-based prototype models.
\item Specifically, we provide an implementation of 22 metrics, including 13 novel contributions, via our open-source library \textit{QuanProto}.
\item Furthermore, we conducted a comprehensive analysis of three prominent part-based prototype models, \ie \textit{ProtoPNet}~\cite{chen2019thislookslikethat}, \textit{ProtoPool}~\cite{rymarczyk2022interpretable} and \textit{PIPNet}~\cite{nauta2023pip}. We provide new insights and demonstrate the utility and applicability of our library.
\item Our evaluation offers a broad perspective on potential application scenarios by utilizing established datasets such as \textit{CUB200}~\cite{wah2011caltech} and \textit{Cars196}~\cite{krause20133d} while also extending to a more practice-oriented Non-IID context with the \textit{NICO}~\cite{he2021towards} dataset and a multi-label classification task focused on the challenge of learning class-independent features from animals using the \textit{AWA2}\cite{xian2018zero} dataset.
\end{enumerate}

\section{Related Work}
This work addresses a broad spectrum of subdomains that are relevant to prototype-based networks. Central to this line of research is the idea of learning human-interpretable concepts, a property particularly important in high-stakes applications. While prototype-based models represent a widely adopted approach in this domain, they also exhibit notable limitations. These shortcomings highlight the need of practical and comprehensive analysis tools.

\paragraph{Concept-based Artificial Intelligence}
Central to concept-based explanation is the notion of conceptualization~\cite{schank1975structure} and concepts which has already been investigated in different fields, most notably in case-based reasoning~\cite{kolodner1992introduction,sormo2005explanation}. The basic idea is to focus on human understandable higher-order concepts not only related to single features. Here, a recent and closely related direction is concept-based artificial intelligence~\cite{poeta2023concept-survey}.    Examples of this approach are \eg concept bottleneck models~\cite{CBMKOH2020} or concept activation vectors~\cite{kim2018interpretability}. Another approach, which shares similarities with prototype-based models, is \textit{TesNet} by Wang et al. \cite{Wang_2021_ICCV_Tesnet}. Here, an interpretable embedding space constructed by basis-concepts is learned.

\paragraph{Architectures}
The prototypical part network (\textit{ProtoPNet}) from Chen et al.~\cite{chen2019thislookslikethat} is one of the pioneering networks that popularized the concept of part prototype networks in image classification.  In addition to the models examined in this article, other networks have been developed that combine the prototype approach with many concepts from the ML field. Nauta et al.~\cite{nauta2021neuraltree} varied the classification process, by replacing the fully connected layer at the end of the model through a decision tree (\textit{ProtoTree})~\cite{nauta2021neuraltree}.   \textit{ProtoPShare} \cite{protopshare} uses an iterative post-processing technique to merge prototypes from a \textit{ProtoPNet} model, effectively creating shared prototypes between classes and reducing the amount of needed prototypes. Xue et al. \cite{xue2022protopformer} adapted the idea of \textit{ProtoPNet} to be used together with vision transformers as feature extractors. Prototype-based deep learning models have also been used for time series~\cite{ming2019interpretablesequence}, graph data~\cite{zhang2022protgnn}, or image segmentation tasks~\cite{ProtoSeg}. There is also work, which combines deep prototype models with an autoencoder architecture~\cite{ProtoVAE}. Other variants of the \textit{ProtoPNet} architecture, include~\cite{pach2024lucidppn},~\cite{van_der_Klis_2023_ICCV_PDiscoNet},~\cite{carmichael2024probably_flow},~\cite{donnelly2022deformable}. Li et al. \cite{li2024overview} provide a comprehensive overview of the design choices and prototype formulations that are used in the literature analysing prototype quality relations.

\paragraph{Application Domain}
Part-based prototype models have been used in several industrial application settings \eg  to inspect power grids visually~\cite{stefenon2022semiprotopnet}, in the classification of MRI scans to detect Alzheimer's disease~\cite{ProtoPNetAlzheimer} or X-Ray images to classify Chest X-rays~\cite{Kim_2021_XProtoNet}. Carlone et al.~\cite{ProtoPNetBreast} propose a prototype-based model to identify breast cancer, highlighting the acceptance of visual explanations.

\paragraph{Limitations} 
Besides the success of  \textit{ProtoPNet} and the improvement through architectural changes, there are also limitations and some critique regarding  prototypical networks. Elhadri et al. \cite{elhadri2025looks} survey general quality issues regarding the interpretability of learned prototypes in these models. Xu-Darme et al.~\cite{xu2023sanitychecksforpatch} criticize the way \textit{ProtoPNet} and \textit{ProtoTree} visualize the learned prototypes and propose to use different saliency methods to overcome this problem. Hoffmann et al.~\cite{hoffmann2021looks} point out that \textit{ProtoPNets} struggle with adversarial examples or compression artifacts in the input. Nevertheless, Hoffmann et al. do not want to dispute part prototypical networks, but raise the awareness of these problems. Saralajew et al. \cite{saralajew2020fast} contribute to this discussion by introducing a method for fast and provable adversarial robustness certification for Nearest Prototype Classifiers (NPCs).

\paragraph{Evaluation of Prototypical-Part Models}
In the literature, there have been several approaches aiming at evaluating different prototypical-part models. These include different evaluation methods as well as specific metrics. The work of Huang et al.~\cite{huang2023evaluation}, \eg proposes two metrics called \textit{consistency score} and \textit{stability score} that focus on the semantic properties of prototypes alongside a new network architecture to improve these metrics. Recently, there has also been work to create frameworks to simplify the utilization of different types of part-based prototype models, \cf \textit{ProtoPNeXt}~\cite{willard2024looksbetterthatbetter} and the work on the library \textit{CaBRNet}~\cite{xu2024cabrnet}. Ma et al.~\cite{ma2024looks}, try to improve the visualization of the prototypes using multiple examples from the training set to describe the concept behind a learned prototype. The framework \textit{HIVE}, created by Kim et al.~\cite{Kim_2021_XProtoNet}, is designed for the evaluation of explanation techniques. They highlight key issues that are crucial when humans must interpret these explanations. Nauta et al.~\cite{nauta2021looks} examine, why the model considers a prototype and a patch of the input image as similar. A comprehensive survey focusing on the interpretability assessment of part-based prototype models is also given by Nauta et al.~\cite{nauta2023co12}.

\section{Datasets}
For evaluating the respective networks in diverse domains, we selected several datasets from the areas of (multi-label) classification, in particular fine-grained prototypical classification, and general object detection in a Non-IID setting.

\paragraph{Caltech-UCSD Birds (CB200) Dataset}
The first fine-grained dataset is Caltech-UCSD Birds-200-2011 (\textit{CUB200}) \cite{wah2011caltech}, that features a challenging collection of bird species. This dataset is frequently used as a benchmark for part-based prototype methods. The rich annotations allow for a detailed evaluation of the importance of individual object parts in the classification process, enabling a comprehensive assessment of the prototypes.  The dataset contains 11,788 images of 200 bird species with segmentation masks, bounding boxes, part locations, and attribute labels.

\paragraph{Stanford Cars Dataset (Cars 196)}
For our second fine-grained dataset, we selected the Stanford Cars dataset (\textit{Cars196}), introduced by Krause et al.~\cite{krause20133d}. Unlike the natural objects in the \textit{CUB200} dataset, this dataset focuses on fine-grained classification of manufactured objects, and is similarly used to benchmark part-based prototype methods. It consists of 16,185 images of 196 car models, covering a wide range of vehicle types including sedans, SUVs, coupes, convertibles, pickups, and hatchbacks.

\paragraph{NICO Dataset}
For general object detection in a Non-IID setting, we used the \textit{NICO} Dataset from He et al.~\cite{he2021towards}. This dataset includes 10 animal classes and 9 vehicle classes, though our focus is on the animal classes. Each class is further divided into 10 context categories representing various environments, attributes, and activities. The dataset contains 12,980 images of animals. Essentially, this Non-IID setting is a more realistic scenario, which is also new for part-based prototypical models. Hence, we want to apply this setting mainly through different background environments in the training and test sets to assess if the networks naturally promote object-focused prototypes. Since this cannot be achieved using the provided context categories (as few represent distinct environments), we use a simple image analysis inspired by Cheng et al. \cite{cheng2001color}, based on the HSV colour space~\cite{HSVHSL2024}, to split the dataset into training/validation and test sets. This approach effectively creates a Non-IID scenario with varying distributions across training and test data. Further details are given in the implementation.

\paragraph{Animals with Attributes (AWA2) Dataset}
The last dataset is Animals with Attributes 2 (\textit{AWA2}) introduced by Xian et al.~\cite{xian2018zero}. This dataset is a benchmark for attribute-based classification, a related model domain to part-based prototype models. The dataset comprises 37,322 images across 50 animal classes, with binary and continuous class attributes. For evaluating the performance of part-based prototype methods in a multi-label classification setting, we focus on the given attributes. The AWA2 dataset includes 85 attributes that are used to describe a class. Since not all attributes describe visible features, we only use a subset, \ie those corresponding to the 49 visible attributes.

\section{Prototype Networks}\label{sec:methods}
\begin{figure*}[t]
    \centering
    \includegraphics[width=\textwidth]{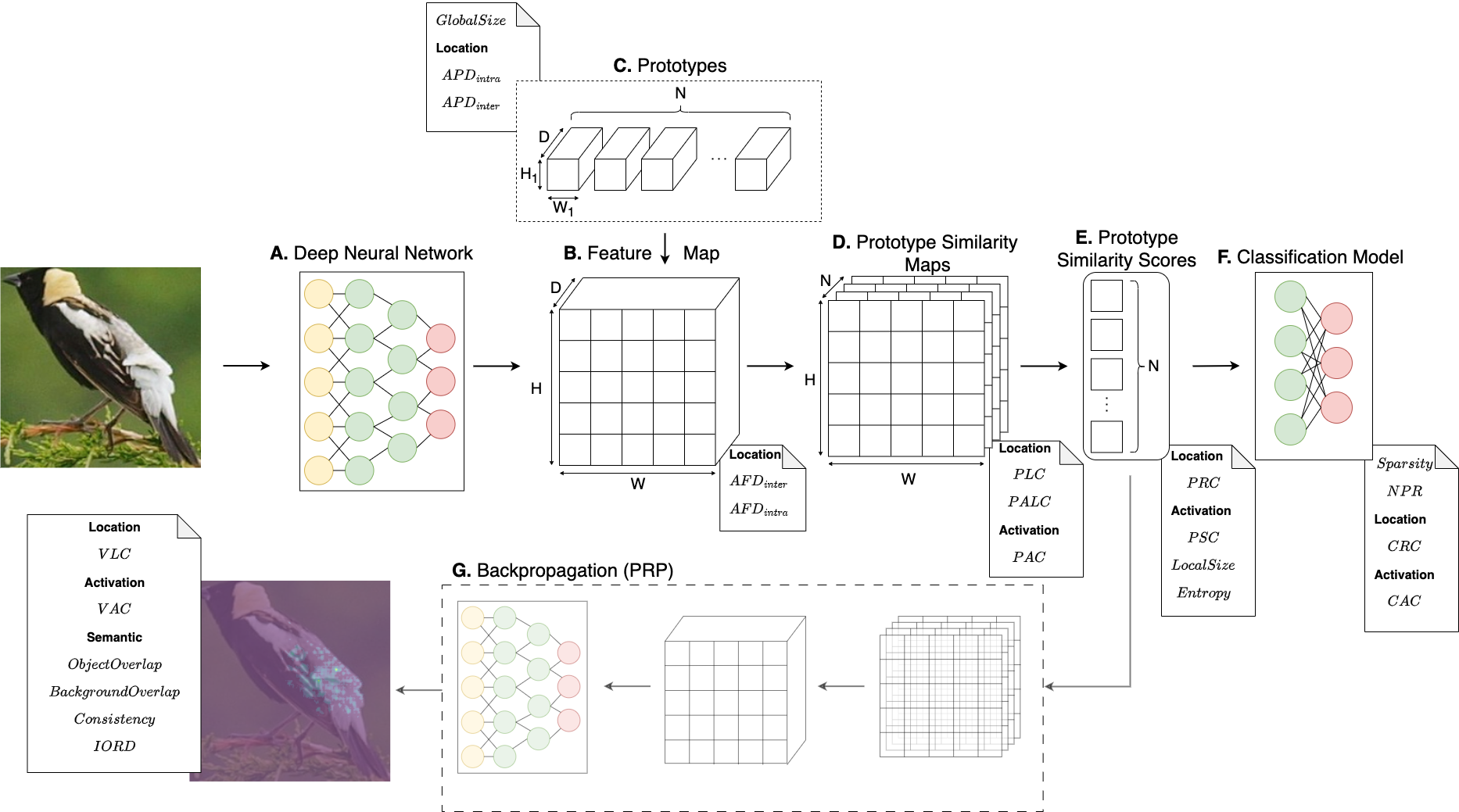}
    \caption{A visual representation of a prototype-based neural network.
    Starting from (A) a deep neural network, such as ResNet, extracting a
    feature map (B), the model identifies the presence of prototypes (C) by
    comparing them with feature vectors; generating similarity maps (D). Classification
    (F) is performed according to the highest similarity scores (E) derived from these maps. Prototypes are visualized (G) using the Prototypical Relevance Propagation (\textit{PRP}) method. Note cards list the metrics that focus on that specific component. Some metrics are used in multiple evaluation techniques, which is later indicated by a subscript, see table \ref{tab:metrics}}
    \label{fig:basic_network}
\end{figure*}
Because this work focuses on a comprehensive set of metrics, we kept our set of models small and selected three prominent networks based on the following criteria: class-specific vs. shared prototypes and explicit vs. indirect prototype representations. We chose \textit{ProtoPNet}~\cite{chen2019thislookslikethat}, the first popular prototype model, that uses class-specific explicit prototype vectors. We further include the \textit{ProtoPool}~\cite{rymarczyk2022interpretable} model as a representative for shared and explicit prototype learning. Finally, we chose \textit{PIPNet}~\cite{nauta2023pip} as a recently introduced architecture for the shared and indirect case. To the best of our knowledge, there is currently no model for the class-specific indirect case. Other models with different classification approaches, such as hierarchical classifications or other backbones such as transformers, are further important studies that should be investigated in connection with the interpretability of prototypes. However, due to a lack of implementations of visual methods and complex adjustments for the multi-label scenario, we decided to start with models that can be easily adapted to our training and evaluation setups.

Each network uses a similar architecture visualized in Figure~\ref{fig:basic_network}, the components are labelled with the letters (A)-(G) in the figure: Starting with a deep neural network (A) as backbone. This network extracts a feature map (B) from the input image. The important characteristic of a part-based prototype model is to calculate the presence of prototypes in the feature map. In the case of a model with specific prototypes, each prototype (C) is compared with the feature vectors in (B) resulting in a set of similarity maps (D). In the case of indirect prototypes, the feature map is further processed and directly interpreted as a collection of similarity maps (D) without an explicit comparison. The final classification (F) is then obtained based on the highest similarity score (E) in each map. Another important characteristic of these models is the visualization of such prototypes (G). In our case, this is done with the Prototypical Relevance Propagation (\textit{PRP}) method that propagates the similarity score back to the input.

Below, we summarize each applied prototype network architecture, highlighting the unique design choices and training procedures.

\paragraph{ProtoPNet}
The first network is \textit{ProtoPNet}. It follows the general architecture shown in Figure~\ref{fig:basic_network} consisting of a convolutional neural network as a backbone followed by two additional $1 \times 1$ convolutional layers to adjust the output dimension $D$. This results in a feature map \(\mathbf{z} \in \mathcal{R}^{H \times W \times D}\) with individual vectors \(\mathbf{\tilde{z}} \in \mathcal{R}^{1 \times 1 \times D}\). The feature map is then processed by a prototype layer, that holds a set of $n$ prototypes \(\mathbf{P}=\left\{\mathbf{p}_m\right\}_{m=1}^n\) with \(\mathbf{p}_m \in \mathcal{R}^{1 \times 1 \times D}\), such that each prototype has the same dimensions as one feature vector. A pre-determined number of prototypes $m_k$ is allocated for each class $k \in$ $\{1, \ldots, K\}$.  The prototype layer computes a set of similarity maps \(\mathbf{A}=\left\{\mathbf{a}_m\right\}_{m=1}^n\) based on the squared L2 distance between a prototype and a feature vector.
\begin{equation}
    \mathbf{a}_m=\Biggl\{\log \left(\frac{\left\|\tilde{\mathbf{z}}-\mathbf{p}_m\right\|_2^2+1}{\left\|\tilde{\mathbf{z}}-\mathbf{p}_m\right\|_2^2+\epsilon}\right)\Bigg| \tilde{\mathbf{z}} \in \mathbf{z} \Biggr\} \in \mathcal{R}^{H \times W \times 1}
\label{eq:proto_similarity}
\end{equation}
The similarity score of a prototype is then calculated using a max pooling operation $\mathbf{s}_m=\max \left(\mathbf{a}_m\right)$ and lastly used in the fully connected layer to make the classification.

\textit{ProtoPNet} uses a multistep training process, separating the training of the last layer from the rest of the network. At the beginning, the last layer is initialized using $1$ for weights connecting a prototype to its corresponding class and $-0.5$ otherwise. This ensures the learning of class specific prototypes. The first training step optimizes the backbone and prototype layer to learn a meaningful latent space, clustering prototypes from the same class and separating them from prototypes of other classes. This is done using a linear combination of a classification loss and two novel losses, the cluster loss and separation loss $\lambda_C \mathcal{L}_C+\lambda_{cl} \mathcal{L}_{cl}+\lambda_{sep} \mathcal{L}_{sep}$. In the second step, each prototype is first projected onto the nearest feature vector from its assigned class. The last layer is then fine-tuned to improve classification performance, using only the classification loss and an L1 regularization term to keep the positive reasoning characteristic. Chen et al.~\cite{chen2019thislookslikethat} observed that some prototypes focus on the background. These prototypes contradict the goal of the model to learn class-specific object parts. To address this, the authors propose a pruning step after training, which removes such prototypes. We use both the original and pruned version in the evaluation to analyse the effectiveness of this approach.

Our experiments include a multi-label dataset. To adapt the training objectives for this setting, we switch from a cross-entropy loss to a multi-label margin loss. Other common loss functions for multi-label classification include binary cross-entropy loss and multi-label soft margin loss. However, these losses tend to learn negative weights in the classification layer for labels not present in the image, which conflicts with the L1 regularization used to promote positive reasoning in the classification layer. The multi-label adaption for the cluster and separation loss is done by calculating the mean cluster and separation loss across all classes $C_{y_i} = \{\,k \mid y_i^k = 1\,\}$, i.e., the set of classes assigned to sample $y_i$.

\begin{equation}
    \mathcal{L}_{cl}=\frac{1}{n} \sum_{i=1}^n \frac{1}{|C_{y_i}|} \sum_{k \in C_{y_i}} \min _{j: \mathbf{p}_j \in \mathbf{P}_k} \min_{\tilde{\mathbf{z}} \in \mathbf{z}} \left\|\tilde{\mathbf{z}}-\mathbf{p}_j\right\|_2^2
\label{eq:cluster_multi}
\end{equation}

\begin{equation}
    \mathcal{L}_{sep}=-\frac{1}{n} \sum_{i=1}^n  \frac{1}{|C_{y_i}|} \sum_{k \in C_{y_i}} \min _{j: \mathbf{p}_j \notin \mathbf{P}_k} \min _{\tilde{\mathbf{z}} \in \mathbf{z}}\left\|\tilde{\mathbf{z}}-\mathbf{p}_j\right\|_2^2
\label{eq:sep_multi}
\end{equation}
We argue that this is a fitting adaption that does not contradict with the goal described in the original study.

\paragraph{ProtoPool}
The \textit{ProtoPool}~\cite{rymarczyk2022interpretable} model by Rymarczyk et al., addresses several limitations of \textit{ProtoPNet}. It introduces a new similarity function  and an automatic, fully differentiable assignment of prototypes to classes, that enables the network to learn shared prototypes.  The model itself follows the architecture of \textit{ProtoPNet} consisting of a CNN backbone with additional convolutional layer to adjust the output shape, a prototype layer, and a fully connected layer. The prototype layer contains a set of $n$ trainable prototypes \(\mathbf{P}=\left\{\mathbf{p}\right\}^n\) with \(\mathbf{p} \in \mathcal{R}^{1 \times 1 \times D}\), similar to \textit{ProtoPNet}. A novel part is the assignment matrix with $L$ prototype slots for each class, where each slot is implemented as a distribution $q_l \in \mathbb{R}^N$ over all available prototypes. The similarity maps \(\mathbf{A}=\left\{\mathbf{a}\right\}^n\) are computed using the same equation \ref{eq:proto_similarity} as in \textit{ProtoPNet}. To enhance prototype locality, the authors introduce a focal similarity function, which computes the difference between the maximum similarity and the average similarity.
\begin{equation}
    g=\max (\mathbf{a})-\operatorname{mean} (\mathbf{a})
\end{equation}
After calculating all similarity scores, the assignment matrix is used to compute an aggregated similarity value $g_l$ for each slot of a class as follows $g_l=$ $\sum_{i=1}^N q_l^i g_i$, which is then used in the fully connected layer to make the final prediction.

\textit{ProtoPool} follows the same multistep training process as \textit{ProtoPNet}. In the first step, the last layer is initialized with a $1$ for a slot assigned to a class and $0$ otherwise, and then fixed during the training of the backbone and prototype layer. The optimization uses classification loss, cluster loss, and separation loss as in \textit{ProtoPNet}. To learn the distributions $q_l$, the authors apply a Gumbel-Softmax layer, that adds additional noise to the distribution at the start of the training process. This noise is then reduced during the training period, so the distribution can converge to a one-hot encoded vector. To prevent that a prototype is assigned to  multiple slots of the same class, an additional orthogonal loss is introduced. Thus, the overall loss term becomes $\lambda_C \mathcal{L}_C+\lambda_{cl} \mathcal{L}_{cl}+\lambda_{sep} \mathcal{L}_{sep}+ \lambda_{o} \mathcal{L}_{o}$.

After this training step the prototypes are projected to the nearest feature vectors similar to \textit{ProtoPNet} and the last layer is fine-tuned to optimize the classification performance.

The adaptation for multi-label training objectives follows the same approach as the \textit{ProtoPNet} model. This involves switching from cross entropy loss to a multi-label margin loss and using the mean cluster and separation loss across all labels as illustrated in equation \ref{eq:cluster_multi} and \ref{eq:sep_multi}. No adaptation is necessary for the orthogonal loss $\mathcal{L}_{o}$, as it computes cosine similarities between each slot without utilizing class information.

\paragraph{PIPNet}
Nauta et al. introduced \textit{PIPNet}~\cite{nauta2023pip}, which incorporates two novel loss functions based on CARL (Consistent Assignment for Representation Learning)~\cite{silva2022representation} to learn prototypes that more accurately align with real object parts. Unlike \textit{ProtoPNet} or \textit{ProtoPool}, \textit{PIPNet} does not use explicit prototype vectors. This removes the need to project prototypes onto nearby latent patches. The architecture includes a CNN backbone (without additional convolutional layers), followed by a softmax layer and a fully connected layer.  The feature map \(\mathbf{z} \in \mathcal{R}^{H \times W \times D}\) from the backbone is treated as $D$ two-dimensional $(H \times W)$ prototype similarity maps. The softmax operation across the $D$ dimension encourages a patch from the feature map $\boldsymbol{\tilde{z}}_{h, w}$, to belong to exactly one prototype similarity map. The similarity scores are again computed via a max-pooling operation and used in the fully connected layer for classification.

Unlike \textit{ProtoPNet} and \textit{ProtoPool}, \textit{PIPNet} trains the entire model without fixing the last layer. To ensure positive reasoning, the fully connected layer is constrained to have only positive weights. Another key distinction is the Siamese data augmentation pipeline, where two views of the same image patch are created. These views are used in the first novel loss function, the alignment loss, that optimizes the two views to belong to the same prototype. This promotes the extraction of more general features. The second loss function, the tanh-loss, ensures that each prototype is represented at least once in a training batch to prevent trivial solutions. The last loss function is the standard classification loss. The overall loss is defined as a linear combination of these  three terms $\lambda_C\mathcal{L}_C + \lambda_A\mathcal{L}_A + \lambda_T\mathcal{L}_T$.

\textit{PIPNet} follows a multistep training process. In the first step, only the backbone is optimized using the alignment and tanh loss to learn prototypes representing general, class-independent features. In the second step, the entire network is optimized using all loss terms, including classification loss. To promote sparsity during training, ensuring that each class uses a minimal number of prototypes for classification, the output scores $o$ are further modified.
\begin{equation}
    \boldsymbol{o}=\log \left(\left(s_m \omega_c\right)^2+1\right)
\end{equation}
Here, $s_m$ are the prototype similarity scores and $\boldsymbol{\omega}_c$ the weights of the fully connected layer. The natural logarithm is employed to promote sparsity, as decreasing the weights has a higher loss gain than increasing weights when the weights become too large.

Since \textit{PIPNet's} loss functions are independent of class assignments, they can be used directly for multi-label classification. The only modification required is that the classification loss $\mathcal{L}_C$ be calculated using the multi label margin loss rather than the cross entropy loss.

\subsection{Prototype Visualization}
The post-hoc visualization method is not bound to the individual models, allowing us to select the most suitable method for the networks. The original implementation of \textit{ProtoPNet} employs a simple model-agnostic upscaling method, which was also used by \textit{ProtoPool} and \textit{PIPNet}. This method upscales the similarity map of a prototype to the size of the input using a cubic interpolation to visualize the similarity pattern of the prototype within the image. A key disadvantage is that it relies solely on the similarity map and assumes the preservation of spatial information in the feature extraction process. There are other model-specific methods that can better visualize the feature extraction process of the model. Based on these arguments and the findings from Xu-Darme et al.\cite{xu2023sanity}, which compared various visualization methods, we selected the Prototypical Relevance Propagation (\textit{PRP}) method from Gautam et al.\cite{gautam2023looks} for visualizing the prototypes (see Figure~\ref{fig:basic_network}).

\section{Evaluation Metrics}\label{sec:metrics}

In this section, we discuss the techniques and metrics used to evaluate the part-based prototype models. Following the Co-12 properties from Nauta et al.~\cite{nauta2023co12} as a guideline, we choose metrics to evaluate the output-completeness, continuity, contrastivity, covariate complexity, and compactness. 

\onecolumn
\begin{table}[h]
\begin{adjustbox}{max width=\linewidth, max height=\textheight}
        \begin{tabular}{ll}
            \toprule
            \toprule
            \multicolumn{2}{l}{\makecell[l]{
\textbf{Output Completeness}: Evaluates whether the visualization covers all pixels relevant to a prototype by
perturbing unhighlighted\\ pixels and measuring the induced changes in the model and explanations.}}
            \\
            \midrule
            $PLC_{out} \downarrow$
            & \makecell[l]{
\textbf{M}easurement of maximum similarity location shift after perturbation. Large shifts indicate poor alignment
\\between the visualization location and the nearest feature location in the feature map.} 
            \\
            $PSC_{out} \%\downarrow$ \cite{sacha2023interpretability}
            & \makecell[l]{
\textbf{Q}uantifies the change in the maximum similarity value. A high change indicates that the relevant image\\
region of the nearest feature is not fully captured by the visualization}
            \\
            $PALC_{out} \%\downarrow$
            & \makecell[l]{
\textbf{M}easures the spatial similarity distribution shift across the feature map. Large deviations indicate
\\poor alignment between visualization and spatial feature extraction.}
            \\
            $PAC_{out} \%\downarrow$ 
            & \makecell[l]{
\textbf{Q}uantifies changes in overall similarity magnitude irrespective of location. High values suggest that
\\the visualization did not cover the image/object characteristics similar to the prototype.} 
            \\
            VLC$\%\downarrow$ \cite{sacha2023interpretability}
            & \makecell[l]{
\textbf{M}easures displacement of the visualization’s highlighted region between original and perturbed images. Large
\\displacements imply a strong impact of the induced spatial feature extraction changes on the visualization method}
            \\
            VAC$\%\downarrow$
            & \makecell[l]{
\textbf{Q}uantifies changes in explanation activation between original and perturbed images. High values reveal
\\a strong impact of the induced information extraction change on the visualization.}
            \\
            \midrule
            \multicolumn{2}{l}{\makecell[l]{
\textbf{Continuity}: Assesses prototype stability under image-wide, low-level perturbations, indicating whether the model has learned\\ robust, high-level semantic features.}} 
            \\
            \midrule
            $PLC_{conti} \downarrow$
            & \makecell[l]{
\textbf{M}easures the shift in the maximum-similarity location after global low-level perturbations. Large shifts imply
\\a misalignment to the feature map interpretation as spatial aligned high level object part representations.} 
            \\
            $PSC_{conti} \%\downarrow$\cite{sacha2023interpretability}
            & \makecell[l]{
\textbf{Q}uantifies changes in the maximum similarity value under global perturbations. High values indicate
\\the model learned unstable prototypes representing low level image characteristics.} 
            \\
            $PALC_{conti} \%\downarrow$
            & \makecell[l]{
\textbf{M}easures changes in the overall spatial similarity pattern after low-level perturbations. Large changes
\\indicate spatial instability in the feature extraction process.} 
            \\
            $PAC_{conti} \%\downarrow$
            & \makecell[l]{
\textbf{Q}uantifies changes in overall similarity magnitudes after low-level perturbations. High values indicate
\\large movement in the embedding space showing features represent low level image characteristics.}
            \\
            $PRC_{conti} \downarrow$ \cite{sacha2023interpretability}*
            & \makecell[l]{
\textbf{R}eports rank changes of the prototype score relative to other prototypes under perturbations. Large rank
\\shifts indicate prototypes are influenced differently, indicating different degrees of semantic representation.} 
            \\
            CAC\% $\downarrow$
            & \makecell[l]{
\textbf{M}easures changes in output logits induced by low-level perturbations. Large variations indicate that nuisance
\\characteristics substantially affect the classification process.} 
            \\
            CRC $\downarrow$
            & \makecell[l]{
\textbf{R}eports rank changes of the predicted class under perturbations. Large rank drops indicate limited class-level
\\robustness and reliance on such non-semantic information.}
            \\
            \midrule
            \multicolumn{2}{l}{\makecell[l]{
\textbf{Contrastivity}: Assesses whether different prototypes capture distinct object properties, promoting robust
classification and
\\informative explanations.}}
            \\
            \midrule
            $PLC_{contra}\uparrow$
            & \makecell[l]{
\textbf{M}easures the location distance between prototype similarity maxima in a sample image. Larger distances
\\ indicate that prototypes focus on distinct object parts.}
            \\
            $PALC_{contra}\%\uparrow$
            & \makecell[l]{
\textbf{M}easures spatial separation of prototype similarity pattern in a sample image. Higher values denote
\\disentangled unique feature vectors resulting in focused prototype activations.} 
            \\
            $APD_{intra} \uparrow$
            & \makecell[l]{
\textbf{C}omputes average pairwise distance between prototypes within a class. Larger distances indicate prototypes
\\represent distinct object parts.} 
            \\
            $AFD_{intra} \uparrow$
            & \makecell[l]{
\textbf{C}omputes average distance between feature vectors closest to a prototype. Large deviations to
\\the average prototype distance indicate a distorted class cluster distribution.} 
            \\
            $APD_{inter} \uparrow$ \cite{wang2023learning}
            & \makecell[l]{
\textbf{M}easures average distance between prototypes across classes. Larger distances indicate well-separated class
\\representations in the embedding space.}
            \\
            $AFD_{inter} \uparrow$ \cite{wang2023learning}
            & \makecell[l]{
\textbf{M}easures average distance between feature vectors closest to a prototype across classes. Small values
\\imply minimal class distinct information preserved in the feature extraction process.} 
            \\
            Entropy $\downarrow$
            & \makecell[l]{
\textbf{C}omputes the entropy of the similarity scores from a prototype across the test set. Lower entropy indicates \\better embedding space  separation with only intended class features near the prototype.} 
            \\
            \midrule
            \multicolumn{2}{l}{\makecell[l]{
\textbf{Covariate Complexity}: Assesses the semantic focus of prototypes by quantifying overlap between prototype visualizations,\\ object masks, and annotated parts.}}
            \\
            \midrule
            ObjectOverlap $\%\downarrow$ \cite{wang2023learning}
            & \makecell[l]{
\textbf{M}easures object mask coverage of visualizations. Lower values indicate part-level focus rather than broad
\\object focus.}
            \\
            BackgroundOverlap $\%\downarrow$
            & \makecell[l]{
\textbf{M}easures the background overlap of visualizations. Low overlap indicates minimal reliance on non-object
\\context.}
            \\
            IORD $\uparrow$ \cite{wang2023learning}*
            & \makecell[l]{
\textbf{M}easures the difference between average activations inside versus outside the object region. Higher scores indicate
\\object-centric explanations rather than background reliance.}
            \\
            Consistency $\%\uparrow$ \cite{huang2023evaluation}*
            & \makecell[l]{
\textbf{M}easures agreement between a prototype’s visual focus and annotated object parts across images. Higher
\\consistency indicates stable and interpretable connection to semantic object parts.}
            \\
            \midrule
            \multicolumn{2}{l}{\makecell[l]{
\textbf{Compactness}: Assesses model size and interpretability of the classification process.
}}
            \\
            \midrule
            Global Size $\downarrow$ \cite{nauta2023pip}
            & \makecell[l]{
\textbf{C}ounts the number of active prototypes used by the model. Smaller values increase interpretability of the
\\classification process by reducing its overall size.} 
            \\
            Sparsity$\%\uparrow$ \cite{nauta2023pip}
            & \makecell[l]{
\textbf{M}easures sparsity of the classification layer. Higher sparsity indicates reliance on fewer prototypes per class,\\ increasing interpretability.}
            \\
            NPR $\downarrow$
            & \makecell[l]{
\textbf{C}omputes the ratio of negative to positive classification weights. Lower ratios improve interpretability by
\\aligning decisions with part-based evidence and not absence.}
            \\
            Local Size $\downarrow$ \cite{nauta2023pip} 
            & \makecell[l]{
\textbf{C}ounts prototypes near the extracted features. Smaller values increase interpretability, meaning only a small
\\ number of prototypes are used for classification.}
            \\
            \bottomrule
            \bottomrule
        \end{tabular}
\end{adjustbox}
        \caption{Overview of metrics by evaluation category with references to original studies. Asterisks (*) indicate metrics inspired by similar measures in the cited work. Up/down arrows denote the direction of improvement; some directions differ across categories depending on the desired outcome.}
        \label{tab:metrics}
\end{table}
\twocolumn
We use a total of 22 metrics, including 13 metrics that are novel to this domain. Figure~\ref{fig:basic_network} provides an overview of which metrics address which model components (A-G). Additional categories indicate whether a metric measures location, activation, or semantic properties. Some metrics are used in multiple evaluation techniques, indicated by a subscript. An overview of which metric is used in which technique, with a brief description of how the metric should be interpreted in each technique, can be found in Table~\ref{tab:metrics}.

We only use the top-5 most activated prototypes of a sample image to evaluate prototype quality-related metrics. This approach is aligned with the actual explanation scheme, where only a few prototypes are visualized to explain a decision to the user. To evaluate the general performance of the networks, we use accuracy, top-3 accuracy, and F1 score.

\subsection{Output-Completeness Evaluation}
The output-completeness evaluation focuses on the visualization method's ability to highlight all relevant parts in the input image. To evaluate the output completeness of the \textit{PRP} method, we consider the study by Sacha et al. \cite{sacha2023interpretability}, which evaluates the spatial misalignment of visualization methods. The introduced metrics measure the location change in the visualization \textit{VLC} and the prototype activation change \textit{PSC} when a perturbation is applied to the input image. We extend this list by including the activation change of the visualization \textit{VAC}, the location change of the maximum activation in the feature map \textit{PLC}, as well as the general location and activation change of the similarity pattern, \textit{PALC} and \textit{PAC}, respectively, thereby achieving comprehensive coverage of the model components, see Figure~\ref{fig:basic_network}. Some metrics from Sacha et al. \cite{sacha2023interpretability} have been renamed for consistency in terminology. Furthermore, some metrics are also used in continuity and contrastivity evaluation with different input variables. Here, the according formulas are only presented once.

Consider an image $x$ and the perturbed counterpart $\overline{x}_i$, created based on the visualization of the prototype $p_i \in P$. Let $v_{p_i}(x)$ be the saliency map of prototype $p_i$. Each saliency map is pre-processed and only contains pixels with a relevance value above the 95th percentile. Let $b_{p_i}(x)$ be the bounding box around this activated region of the saliency map. For the output-completeness evaluation, we perturb an image $x$ to $\overline{x}_i$ by adding a Gaussian noise mask with a standard deviation of $0.05$ to the pixels around the bounding box; see Figure~\ref{fig:completeness_complexity} for a process visualization. Further, let $a_{p_i}(x)$ be the similarity map and $s_{p_i}(x)$ the similarity score of the prototype. 

If the visualization method correctly highlights the relevant parts of the prototype, then we expect minimal changes in the model components (see Figure~\ref{fig:basic_network}), resulting in low scores for these metrics.

The visualization location change \textit{VLC}, previously introduced as the prototypical part location change by Sacha et al. \cite{sacha2023interpretability}, assesses the change in the location of the bounding box following a perturbation by calculating the intersection over union.
\begin{equation}
    \mathit{VLC}=1- \frac{\left|b_{p_i}(x) \cap b_{p_i}\left(\overline{x}_i\right)\right|}{\left|b_{p_i}(x) \cup b_{p_i}\left(\overline{x}_i\right)\right|}
\end{equation}
To evaluate the similarity change, we use the prototype similarity change \textit{PSC}, formerly introduced as prototype activation change by Sacha et al. \cite{sacha2023interpretability}. This metric computes the difference between two similarity values and is used to measure the relative change in the similarity score after the perturbation is applied.
\begin{equation}
    \mathit{PSC}_{out}= \frac{|(s_{p_i}(x)-s_{p_i}\left(\overline{x}_i\right)|}{s_{p_i}(x)}
\end{equation}
To complete the evaluation of changes in the saliency map, we introduce the visualization activation change \textit{VAC}, which measures the change in relevance scores between two saliency maps. To be invariant to location changes in the visualization, we flatten and sort the saliency maps to get a relevance curve, represented as $\tilde{v}_{p_i}(x)$. The difference in activation is measured by calculating the intersection over union of the two activation curves in the following way. Let $h \times w$ represent the height and width of the saliency map.
\begin{equation}
    \mathit{VAC}=1- \frac{\sum_{j=1}^{h \times w}\min(\tilde{v}_{p_i}(x)_j, \tilde{v}_{p_i}(\overline{x}_i)_j)}{\sum_{j=1}^{h \times w}\max( \tilde{v}_{p_i}(x))_j, \tilde{v}_{p_i}(\overline{x}_i)_j)}
\end{equation}
We also assess location and activation changes in the feature space using the similarity map. In order to measure the location change, we introduce two metrics: First, the prototype location change \textit{PLC}, measuring the distance between two maximum activation locations using the Manhattan distance. The output-completeness evaluation uses the similarity maps of a prototype from the original and perturbed image version.
\begin{equation}
    \mathit{PLC}_{out}= ||\operatorname{argmax}\left(a_{p_i}(x)\right)-\operatorname{argmax}\left(a_{p_i}\left(\overline{x}_i\right)\right)||_1
\end{equation}
The second metric is the prototype activation location change \textit{PALC}. This metric measures the location difference of the hole activation pattern between two similarity maps, analogous to the \textit{VLC} metric. We apply a min-max normalization and use a threshold of $0.5$ to obtain a binary activation pattern, expressed by the $\operatorname{bin}()$ operation. Again, the output-completeness evaluation uses the original and perturbed images to generate the two similarity maps.
\begin{equation}
    \mathit{PALC}_{out}=1- \frac{\left|\operatorname{bin}(a_{p_i}(x)) \cap \operatorname{bin}(a_{p_i}\left(\overline{x}_i\right))\right|}{\left|\operatorname{bin}(a_{p_i}(x)) \cup \operatorname{bin}(a_{p_i}\left(\overline{x}_i\right))\right|}
\end{equation}
In order to evaluate the activation change in similarity maps, we introduce the prototype activation change \textit{PAC}, which operates similar to the \textit{VAC} metric and measures the activation difference between two similarity maps. Let $m \times n$ represent the height and width of the feature map, then
\begin{equation}
    \mathit{PAC}_{out}=1- \frac{\sum_{j=1}^{m \times n}\min(a_{p_i}(x)_j, a_{p_i}(\overline{x}_i)_j)}{\sum_{j=1}^{m \times n}\max( a_{p_i}(x)_j, a_{p_i}(\overline{x}_i)_j)}
\end{equation}
Using these additional metrics, we evaluate the changes in location and activation at each step before the final classification in the general architecture of prototype models, illustrated in Figure~\ref{fig:basic_network}.

\subsection{Continuity Evaluation}
\begin{figure*}[t]
    \centering
        \includegraphics[width=\linewidth]{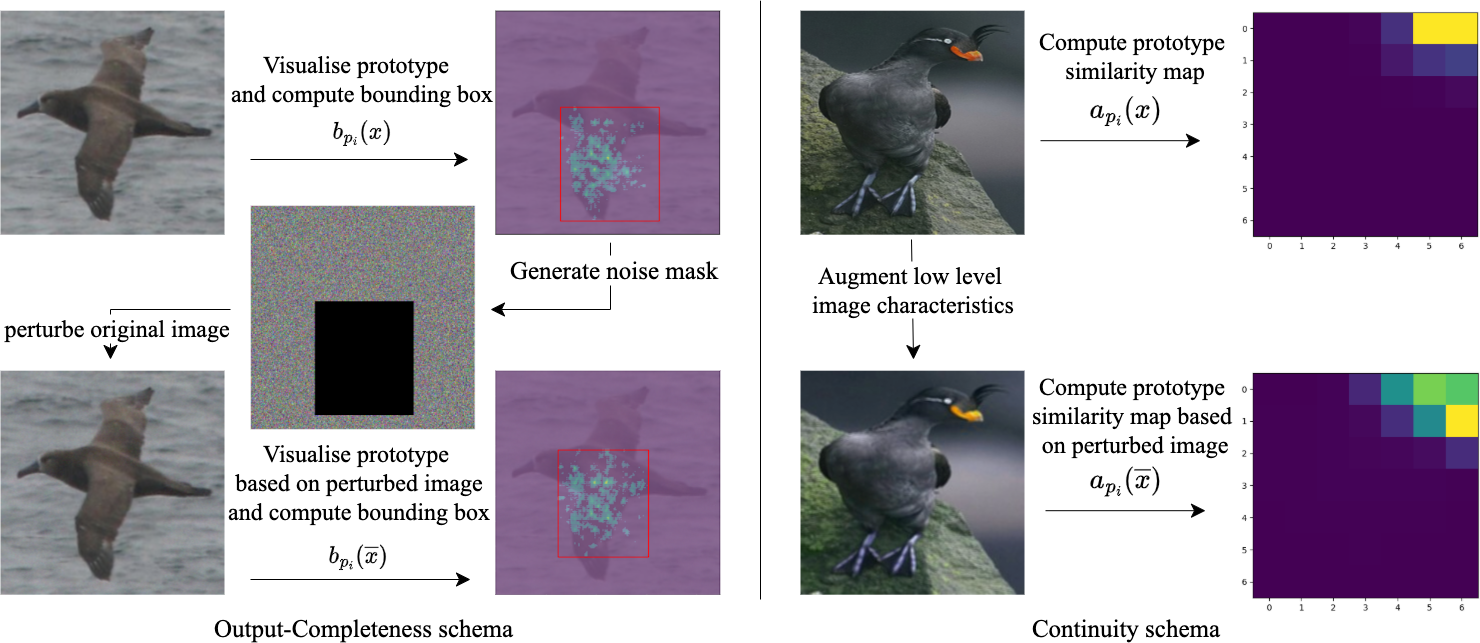}
    \caption{The output-completeness of the visualization method is assessed by measuring the change in different parts of the model when the image is perturbed based on a prototype's visualization. The continuity of a model is assessed by measuring the effect of augmentations to prototypes.}
    \label{fig:completeness_continuity_vis}
\end{figure*}
The continuity evaluation assesses the stability of prototypes under different augmentations. We focus on the study by Hoffmann et al. \cite{hoffmann2021looks}, which assesses prototype stability under noise and compression artifacts, and the studies by Rymarczyk et al. \cite{protopshare} and Nauta et al.~\cite{nauta2021looks}, which examine the influence of photometric augmentations.  We train our model using geometric and photometric augmentations to promote continuity during training, as recommended by Nauta et al.~\cite{nauta2023co12}.  In the evaluation, we perturb the input image $x$ to $\overline{x}$ with photometric and noise augmentations. The individual augmentations are applied all at once with the following settings: brightness $+12.5\%$, contrast $+12.5\%$, saturation $+12.5\%$, hue $0.05$, noise $5\%$, JPEG quality $90\%$, and blur using a $3 \times3$ kernel. A visual presentation of the process is illustrated in Figure~\ref{fig:basic_network}.

We utilize the metrics from the output-completeness evaluation to evaluate the changes in location and activation within the major components of the models (Figure~\ref{fig:basic_network}). This is done by switching the prototype-specific perturbation $\overline{x}_i$ with the general perturbation $\overline{x}$ based on the specified augmentation techniques. The metrics adopted are \textit{PLC}, \textit{PSC}, \textit{PALC}, \textit{PRC}, and \textit{PAC}. The \textit{VLC} and \textit{VAC} scores are excluded, as they focus on assessing the visualization method instead of a prototype. In addition, we use the prototype rank change \textit{PRC} from Sacha et al. \cite{sacha2023interpretability}. This metric measures the prototype's position change in the similarity score ranking of all prototypes.
\begin{equation}
    \mathit{PRC}=\left|r_{p_i}\left(\bar{x}_i\right)-r_{p_i}(x)\right|
\end{equation}
We also introduce the following new metrics. The classification activation change \textit{CAC} measures the change in prediction. Let $K$ be the number of classes and $o(x)$ be the output vector of the model for an image $x$.
\begin{equation}
    \mathit{CAC}=1- \frac{\sum_{j=1}^{K}\min(o(x)_j,o(\overline{x})_j)}{\sum_{j=1}^{K}\max(o(x)_j,o(\overline{x})_j)}
\end{equation}
The other metric is the classification rank change \textit{CRC} that measures the change in the rank of the predicted class. Let $r_c(x)$ be the rank of the predicted class $c$ for the image $x$:
\begin{equation}
    \mathit{CRC}=|r_c\left(\bar{x}\right)-r_c(x)|
\end{equation}

\subsection{Contrastivity}
The contrastivity evaluation is used to assess the differences between prototypes. To evaluate the contrastivity of prototypes, we refer to the study by Wang et al.~\cite{wang2023learning}, which evaluated the contrastivity of prototypes in the feature space. We follow the recommendations of Nauta et al.~\cite{nauta2023co12} to also assess location differences.

Wang et al.~\cite{wang2023learning} measured the average cosine distance between prototypes $\mathit{APD_{inter}}$ and feature vectors $\mathit{AFD_{inter}}$ of different classes. These metrics are designed for networks with class-specific prototypes, so each class has a subset of prototypes $P_k \subset P$, where $P$ is the set of all prototypes and $k$ is the class index. To ensure compatibility with models that do not assign prototypes to classes, we use the ground truth class of the image $x$ to create the subset $P_x$, consisting of the top-5 most activated prototypes in the image. The subset $P_k$ is then the combination of all subsets $P_x$ in the test set $X_{test}$. To create the corresponding feature vector sets $F_k$, we use the nearest feature vectors of the prototypes in the subset $P_x$.

We use the average inter-class distance metrics from Wang et al. \cite{wang2023learning} and also employ the average intra-class distance case $\mathit{APD_{intra}}$ and $\mathit{AFD_{intra}}$ respectively. Following the recommendations from Nauta et al.~\cite{nauta2023co12}, we use the \textit{PLC} and \textit{PALC} metrics to assess the contrastivity of prototype locations in the feature map and introduce a new metric to evaluate the activation discriminativeness of prototypes, achieving a comprehensive model coverage (see Figure~\ref{fig:basic_network}). To use the \textit{PLC} and the \textit{PALC} described in the output-completeness evaluation in this context, we change the variable $a_{p_i}(\overline{x}_i)$ to $a_{p_j \in P_{x}/p_i}(x)$ in order to compare the similarity maps between activated prototypes in image $x$

The original study from Wang et al.~\cite{wang2023learning} does not state the specific formula of the metrics, so the following calculations are our interpretation. The average inter-class prototype distance $\mathit{APD_{inter}}$ from Wang et al. ~\cite{wang2023learning} is the average cosine distance between prototypes of different classes.
\begin{align}
    \mathit{APD_{inter}}&=\frac{1}{K}\sum_{k=1}^{K} \frac{1}{|P_k|}\sum_{i: p_i \in P_k} \frac{1}{|P \setminus P_k|}
    \nonumber \\
    &\cdot \sum_{j: p_j \in P \setminus P_k} \left(1- \frac{p_i \cdot p_j}{||p_i|| \cdot ||p_j||}\right)
\end{align}
We calculate the average inter-class feature distance $\mathit{AFD_{inter}}$ from Wang et al.~\cite{wang2023learning} in the same way using the feature vector set $F$ instead of the prototype set $P$. To evaluate the contrastivity between prototypes from the same class, we introduce the average intra-class prototype distance $\mathit{APD_{intra}}$. The metric measures the average cosine distance of a prototype to all other  prototypes of the same class.
\begin{align}
    \mathit{APD_{intra}}&=\frac{1}{K}\sum_{k=1}^{K} \frac{1}{|P_k|}\sum_{i: p_i \in P_k} \frac{1}{|P_k \setminus p_i|}
    \nonumber \\
    & \cdot \sum_{j: p_j \in P_k \setminus p_i} \left(1- \frac{p_i \cdot p_j}{||p_i|| \cdot ||p_j||}\right)
\end{align}
We also extend this metric to the feature vector case via introducing the average intra-class feature distance $\mathit{AFD_{intra}}$, by switching the prototype set $P$ with the feature vector set $F$. In order to measure the class discriminativeness of a prototype, we compute the Shannon entropy of the similarity scores. Let $S_{p_i}$ be the set of similarity scores of a prototype over all test images in $X_{test}$. We use a max normalization to normalize the similarity scores and then calculate the histogram $hist(S_{p_i})$ of the similarity scores with $U=10$ bins.
\begin{equation}
    \mathit{Entropy}=-\sum_{u=1}^{U} hist(S_{p_i})_u \cdot \log(hist(S_{p_i})_u) 
\end{equation} 
A high entropy indicates that the prototype has no discriminative activation pattern.

\subsection{Covariate Complexity}
The covariate complexity evaluation assesses the complexity of a prototype regarding the interpretability. To evaluate the covariate complexity of prototypes, we focus on the study by Wang et al.~\cite{wang2023learning}, which evaluated the overlap of prototype visualizations with object masks, and the study by Huang et al. ~\cite{huang2023evaluation}, which evaluated how consistently object parts are represented by a prototype.

The following metrics from Wang et al.~\cite{wang2023learning} are again our interpretation, as the study does not provide specific formulas. To determine the activated region of a prototype in the input image, we again use the 95th percentile approximation on the prototype visualization for these metrics.

To assess the overlap between prototypes and objects, Wang et al. \cite{wang2023learning} introduces the content heatmap metric. We renamed this metric to $\mathit{Object\ Overlap}$ to be consistent with our naming scheme. The metric calculates the intersection of the prototype visualization and the object mask. Let $m(x)$ be the object mask of image $x$.
\begin{equation}
    \mathit{Object\ Overlap}=\frac{\left|v_{p_i}(x) \cap m(x)\right|}{\left|m(x)\right|}
\end{equation}
We also adapted the idea of the outside-inside relevance ratio measure from Wang et al.~\cite{wang2023learning}, introducing the inside-outside relevance difference \textit{IORD} as the signed difference between mean inside and outside activation. The adoption of the original metric was made to increase the stability of the measurements. Each saliency map is normalized before the calculation to ensure that \textit{IORD} scores are consistent between different saliency maps.
\begin{align}
\mathit{IORD}&=\frac{\sum_{j=1}^{h \times w} \tilde{v}_{p_i}(x)_j \cdot m(x)_j}{\sum_{j=1}^{h \times w} \mathbb{I}((\tilde{v}_{p_i}(x)_j \cdot m(x)_j)> 0 )}
    \nonumber \\
    & - \frac{\sum_{j=1}^{h \times w} \tilde{v}_{p_i}(x)_j \cdot (1 - m(x)_j)}{\sum_{j=1}^{h \times w} \mathbb{I}((\tilde{v}_{p_i}(x)_j \cdot (1 - m(x)_j))> 0 )}
\end{align}
Here, $\mathbb{I}$ is the indicator function that returns 1 if the condition is true and 0 otherwise.

In order to measure the alignment of prototypes with specific object parts, we adapted the consistency score from Huang et al.~\cite{huang2023evaluation}.  The coverage of an object part is calculated by computing a histogram of the object parts covered by a prototype over the test set $X_{test}$.  This is done by computing the bounding box and adding all object parts within the bounding box to the histogram of that prototype. After processing every image in the test set, we normalize the histograms. This results in a vector with object part percentages, denoted as $l_{p_i}$. The consistency of the prototype is the average coverage measure over the vector $l_{p_i}$.
\begin{equation}
    \mathit{Consistency}=  \frac{1}{|l_{p_i}|} \sum_{j}^{|l_{p_i}|} {l_{p_i}}_j
\end{equation}
In addition, we introduce the $\mathit{Background\ Overlap}$ metric as a counterpart to the $\mathit{Object\ Overlap}$. This metric assesses the area of the activated region in the saliency map that does not overlap with the object.
\begin{equation}
    \mathit{Background\ Overlap}= 1 - \frac{\left|v_{p_i}(x) \cap m(x)\right|}{\left|v_{p_i}(x)\right|}
\end{equation}

\subsection{Compactness}
The compactness evaluation assesses general characteristics of the model. This evaluation will focus on the study by Nauta et al.~\cite{nauta2023pip}, which evaluated global size, local size, and the sparsity of the classification layer. Additionally, we will include the negative-positive reasoning ratio metric in our compactness evaluation.

The global explanation size from Nauta et al.~\cite{nauta2023pip} counts all prototypes with at least one non-zero weight in the classification layer. This definition is not directly applicable to the \textit{ProtoPool} model. Therefore, we will count non-zero weights in the prototype presence matrix from \textit{ProtoPool}. The local explanation size, as described by Nauta et al. \cite{nauta2023pip}, can be defined as follows. Let $\hat{s}_{p_i}(x)$ be the normalized similarity score of a prototype $p_i$ for an image $x$, with $s(x)$ being the similarity score vector of all prototypes for the image $x$.
\begin{equation}
\hat{s}_{p_i}(x)=\frac{s_{p_i}(x)}{\max s(x)}
\end{equation}
Then, we count the number of prototypes with a normalized similarity score above a threshold $\mu$, which is set to $0.1$.
\begin{equation}
\mathit{Local\ Size}=\sum_{i=1}^N \mathbb{I}\left(\hat{s}_{p_i}(x)>0.1\right)
\end{equation}
with $N$ as the total number of prototypes.

The classification Sparsity from Nauta et al.~\cite{nauta2023pip} is the ratio of zero weights in the classification layer. Let $W$ represent the weights in the classification layer, and $\epsilon=0.001$ be the threshold for considering a weight as non-zero to improve the numerical stability. 
\begin{equation}
\mathit{Sparsity}=\frac{|W|-\sum_{w \in W} \mathbb{I}(w>\epsilon \vee w<-\epsilon)}{|W|}
\end{equation}
Here, $|W|$ denotes the total number of weights and the sum counts the number of positive and negative weights that exceed the threshold $\epsilon$.

In order to evaluate the positive reasoning property of the model, we introduce the negative-positive reasoning ratio \textit{NPR}, that calculates the ratio between the number of positive and negative weights in the classification layer. We will again use a threshold of $\epsilon=0.001$ for considering a weight as positive or negative.
\begin{equation}
\mathit{NPR}=\frac{\sum_{w \in W} \mathbb{I}(w<-\epsilon)}{\sum_{w \in W} \mathbb{I}(w>\epsilon)}
\end{equation}
The numerator $\sum_{w \in W} \mathbb{I}(w<-\epsilon)$ counts the number of negative weights that are less than $-\epsilon$, and the denominator $\sum_{w \in W} \mathbb{I}(w>\epsilon)$ counts the number of positive weights that are greater than $\epsilon$.

\section{Experimental Setup}
In this section, we briefly describe the chosen model parameters and architectural settings, as well as the dataset and training setup.

In the original studies, datasets were split into 50\% training and 50\% test data. We will follow a different method based on the study by Raschka et al. \cite{raschka2018model}, which argues that using the same data subset for model selection and final evaluation can result in overly optimistic outcomes. Following these guidelines, we first divide the datasets into 70\% training and 30\% test sets. We then apply 4-fold stratified cross-validation to split the training set into training and validation subsets, maintaining a consistent class distribution across all sets. This results in four different sets of training and validation pairs, with a 52.5\% training and 17.5\% validation split.

We selected the \textit{ResNet-50} architecture with pretrained ImageNet weights as a backbone for all models. The feature maps' spatial dimensions are $7 \times 7$. In the original study from \textit{PIPNet} the dimension is modified to $28 \times 28$. However, it is not described as a core architecture design, so we argue that the use of a uniform dimension setting makes the comparison fairer. For the dimension of explicit prototype vectors in the \textit{ProtoPNet} and \textit{ProtoPool} models, we chose $1 \times 1 \times 128$. The last layer of \textit{ProtoPNet} is initialized with a 1 indicating a prototype belongs to a class and 0 otherwise instead of $-0.5$, because the sparsity objective described by Chen et al.~\cite{chen2019thislookslikethat} could otherwise not be achieved in our experiments. \textit{PIPNet} uses 2048 prototypes for all datasets. \textit{ProtoPNet} uses 2000, 1960, 50 and 490 for \textit{CUB200}, \textit{Cars196}, \textit{NICO} and \textit{AWA2} respectively. \textit{ProtoPool} uses 205, 201, 50, 168 prototypes for the datasets, respectively.

We used an online augmentation setting with the Albumentations library \cite{info11020125}, that supports augmenting bounding boxes, segmentation masks, and key-points along with images, which is crucial for certain metrics in our evaluation. All networks were trained with geometric and photometric augmentations. The original studies from \textit{ProtoPNet} and \textit{ProtoPool} only used geometric augmentations, but due to our  consistency evaluation, we choose to add additional photometric augmentations to promote better results. In the original study, the \textit{PIPNet} model used photometric augmentation for the contrastive learning approach, which is also included in our experiments. The experiments on the \textit{CUB200} and \textit{Cars196} were done on cropped images using the bounding box information.

\begin{table*}[t]
    \resizebox{\linewidth}{!}{
        \begin{tabular}{llllllllll}
            \toprule
            \toprule
            CUB200 & Accuracy$\%\uparrow$ & top-3 Acc.$\%\uparrow$  & F1 score$\%\uparrow$ & Global Size $\downarrow$ & Sparsity$\%\uparrow$ & NPR $\downarrow$ & Local Size $\downarrow$ \\
            \midrule
            ProtoPNet & 70.18 $\pm$ 1.32 & 85.29 $\pm$ 0.73 & 70.39 $\pm$ 1.23 & 2000.00 $\pm$ 0.00 & 99.30 $\pm$ 0.18 & 0.25 $\pm$ 0.19 & 1348.02 $\pm$ 146.53 \\
            ProtoPNet P & 69.12 $\pm$ 1.81 & 85.00 $\pm$ 0.95 & 69.20 $\pm$ 1.60 & 1891.75 $\pm$ 16.40 & 99.10 $\pm$ 0.38 & 0.19 $\pm$ 0.13 & 1261.84 $\pm$ 131.89 \\ 
            ProtoPool & 67.33 $\pm$ 0.24 & 80.42 $\pm$ 0.96 & 67.81 $\pm$ 0.44 & \textbf{205.00} $\pm$ 0.00 & 96.04 $\pm$ 0.60 & 0.54 $\pm$ 0.06 & 168.33 $\pm$ 14.58 \\
            PIPNet & \textbf{74.00} $\pm$ 0.64 & \textbf{86.77} $\pm$ 0.32 & \textbf{73.85} $\pm$ 0.56 & 858.00 $\pm$ 76.80 & \textbf{99.31} $\pm$ 0.09 & \textbf{0.00} $\pm$ 0.00 & \textbf{67.69} $\pm$ 0.30 \\
            \bottomrule
            \bottomrule
            Cars196 & Accuracy$\%\uparrow$ & top-3 Acc.$\%\uparrow$ & F1 score$\%\uparrow$ & Global Size $\downarrow$ & Sparsity$\%\uparrow$ & NPR $\downarrow$ & Local Size $\downarrow$ \\
            \midrule
            ProtoPNet & 82.60 $\pm$ 1.07 & 94.01 $\pm$ 0.56 & 82.63 $\pm$ 0.99 & 1960.00 $\pm$ 0.00 & 99.35 $\pm$ 0.07 & 0.10 $\pm$ 0.04 & 1197.61 $\pm$ 106.00 \\
            ProtoPNet P & 81.77 $\pm$ 0.55 & 93.15 $\pm$ 0.66 & 81.67 $\pm$ 0.46 & 1785.50 $\pm$ 22.78 & 98.78 $\pm$ 0.51 & 0.15 $\pm$ 0.05 & 1080.18 $\pm$ 84.72 \\
            ProtoPool & 83.25 $\pm$ 0.57 & 91.62 $\pm$ 0.42 & 83.29 $\pm$ 0.59 & \textbf{201.00} $\pm$ 0.00 & 99.28 $\pm$ 0.12 & 0.11 $\pm$ 0.07 & 170.47 $\pm$ 51.32 \\
            PIPNet & \textbf{85.78} $\pm$ 0.33 & \textbf{94.68} $\pm$ 0.25 & \textbf{85.73} $\pm$ 0.32 & 514.00 $\pm$ 18.46 & \textbf{99.51} $\pm$ 0.04 & \textbf{0.00} $\pm$ 0.00 & \textbf{68.95} $\pm$ 0.28 \\
            \bottomrule
            \bottomrule
            NICO & Accuracy$\%\uparrow$ & top-3 Acc.$\%\uparrow$ & F1 score$\%\uparrow$& Global Size $\downarrow$ & Sparsity$\%\uparrow$ & NPR $\downarrow$ & Local Size $\downarrow$ \\
            \midrule
            ProtoPNet & 90.04 $\pm$ 1.30 & 96.74 $\pm$ 0.61 & 90.01 $\pm$ 1.32 & \textbf{50.00} $\pm$ 0.00 & 60.25 $\pm$ 8.46 & 0.46 $\pm$ 0.12 & 15.59 $\pm$ 3.83 \\
            ProtoPNet P & 90.19 $\pm$ 1.15 & 96.77 $\pm$ 0.60 & 90.15 $\pm$ 1.17 & \textbf{50.00} $\pm$ 0.00 & 55.05 $\pm$ 15.82 & 0.51 $\pm$ 0.14 & 15.69 $\pm$ 3.83 \\
            ProtoPool & 89.89 $\pm$ 0.92 & 95.94 $\pm$ 0.87 & 89.87 $\pm$ 0.93 & \textbf{50.00} $\pm$ 0.00 & 89.55 $\pm$ 0.66 & 0.04 $\pm$ 0.07 & \textbf{3.38} $\pm$ 0.83 \\
            PIPNet & \textbf{91.40} $\pm$ 0.28 & \textbf{97.49} $\pm$ 0.39 & \textbf{91.49} $\pm$ 0.30 & 214.00 $\pm$ 136.81 & \textbf{98.90} $\pm$ 0.93 & \textbf{0.00} $\pm$ 0.00 & 70.80 $\pm$ 0.65 \\
            \bottomrule
            \bottomrule
            AwA2 & Accuracy$\%\uparrow$ &  top-3 Acc.$\%\uparrow$ & F1 score$\%\uparrow$&Global Size $\downarrow$ & Sparsity$\%\uparrow$ & NPR $\downarrow$ & Local Size $\downarrow$ \\
            \midrule
            ProtoPNet & 43.86 $\pm$ 3.01    & - & 91.74 $\pm$ 0.60 & 490.00 $\pm$ 0.00 & 18.27 $\pm$ 0.95 & 0.95 $\pm$ 0.02 & 184.20 $\pm$ 17.85 \\
            ProtoPNet P & 40.61 $\pm$ 8.45  & - & 91.75 $\pm$ 0.47 & 467.75 $\pm$ 7.32 & 18.16 $\pm$ 0.67 & 1.05 $\pm$ 0.13 & 185.56 $\pm$ 25.42 \\
            ProtoPool & 46.33 $\pm$ 6.45    & - & 91.98 $\pm$ 1.29 & \textbf{168.00} $\pm$ 0.00 & 96.43 $\pm$ 1.30 & 0.24 $\pm$ 0.29 & 87.76 $\pm$ 20.57 \\
            PIPNet & \textbf{62.41} $\pm$ 2.72       & - & \textbf{94.47} $\pm$ 0.56 & 206.00 $\pm$ 10.21 & \textbf{98.68} $\pm$ 0.09 & \textbf{0.00} $\pm$ 0.00 & \textbf{70.38} $\pm$ 0.27 \\ 
            \bottomrule
            \bottomrule
            \end{tabular}
    }
    \caption{General performance and Compactness evaluation results. The results are averaged over 4 runs with standard
    deviation. Training and validation subsets were created using 4-fold stratified cross-validation.}
    \label{tab:general_compact}
\end{table*}
Due to the differences between our training setup and the original studies, we adjusted the respective learning rates, training epochs, and schedulers for each model and dataset. The new parameters were chosen based on hyperparameter tuning with 50 trails for each dataset. To reduce complexity and the computational load, we used a fixed number of 100 joint epochs to tune all models. The individual loss weights were kept the same as the original paper suggested over all datasets. All networks were re-implemented based on the original code for our experiments and are also part of our open-source library.  We point out that our training setup deviates from the optimal learning procedure introduced in the original studies, and we expect a reduction in performance. Therefore, our focus is to evaluate if interpretable properties discussed in the original studies remain in suboptimal learning conditions.

\section{Results}

The general performance, compactness, contrastivity, and continuity evaluations are conducted on all datasets. The output-completeness and complexity evaluations are conducted exclusively on the \textit{CUB200} dataset. The output-completeness evaluation assesses the employed visualization method, in our case the computationally intensive \textit{PRP} method; thus, this method was evaluated only on the \textit{CUB200} dataset. The complexity evaluation is limited to the \textit{CUB200} dataset due to the absence of object masks and part annotations in the other datasets.

The used networks are \textit{ProtoPNet}~\cite{chen2019thislookslikethat}, \textit{ProtoPNet Pruned}~\cite{chen2019thislookslikethat}, \textit{ProtoPool}~\cite{rymarczyk2022interpretable}, and \textit{PIPNet}~\cite{nauta2023pip}. The pruned version of \textit{ProtoPNet} was created using the pruning strategy stated in the original paper.

\subsection{General and Compactness}
Table~\ref{tab:general_compact} presents the outcomes of our evaluations on general performance and compactness. 

The pruned version of \textit{ProtoPNet} exhibits a performance drop compared to the original model. This indicates that prototypes not clearly associated with a single class, presumably focusing on background regions, still play a significant role in the classification process. In other words, classification in \textit{ProtoPNet} relies on a complex and sensitive interplay between prototypes, including those that are not class-distinct. Evidence from the compactness analysis supports this interpretation. First, the \textit{Local Size} shows that a relatively high number of prototypes are active per sample. Second, the \textit{NPR} score improves after removing background prototypes, suggesting that even though the fine-tuning stage was initialized with zeros (instead of the original –0.5), which should bias towards positive reasoning, the model nonetheless learns intricate negative relations during fine-tuning. Interestingly, sparsity often decreases slightly after pruning, which is counterintuitive. One would expect that the targeted background prototypes would be shared across more classes; thus their removal should increase sparsity. Combined with the minor reduction in global size, these findings suggest that pruning does not simply eliminate “background prototypes” but is strongly influenced by the quality of learned prototypes.

On the \textit{CUB200} dataset, \textit{ProtoPool} struggles to discover meaningful shared prototypes compared to its performance on other datasets, such as \textit{Cars196}. This is reflected both in performance rankings and compactness measures. Specifically, \textit{ProtoPool} on \textit{CUB200} exhibits a high proportion of negative weights, low sparsity, and a large number of prototypes active per sample, all indicators of a complex classification process. In contrast, when applied to \textit{NICO} and \textit{AWA2}, \textit{ProtoPool} shows improvements in interpretability over \textit{ProtoPNet}. The \textit{Local Size} is reduced on \textit{NICO}, while \textit{NPR} and \textit{Sparsity} scores improve on \textit{AWA2}. These results suggest that shared prototypes may be more advantageous in multi-label settings than in fine-grained single-label classification tasks.

\textit{PIPNet} achieves the best overall results in both predictive performance and compactness. This illustrates the potential of contrastive learning for prototype-based methods. The outcome is intuitive, as contrastive learning naturally aligns with the goal of learning a high-level semantic prototype representation. Furthermore, the hard constraint imposed on the classification layer does not negatively affect performance, indicating that \textit{PIPNet} does not require a separate fine-tuning phase. In contrast, other models that rely on fine-tuning often learn additional, more complex, relationships to increase predictive performance in this phase. \textit{PIPNet} is also the only model capable of reducing \textit{Global Size} during training without the need for external pruning strategies. This highlights the effectiveness of the softmax-based constraint in limiting the number of active prototypes, even while the tanh loss encourages all prototypes to be active at some point during training. Moreover, the consistently small \textit{Local Size} relative to \textit{Global Size} indicates that this design robustly limits prototype usage per sample, thereby enhancing interpretability without sacrificing accuracy.

\subsection{Contrastivity}
\begin{figure*}[t]
    \centering
    \includegraphics[width=\linewidth]{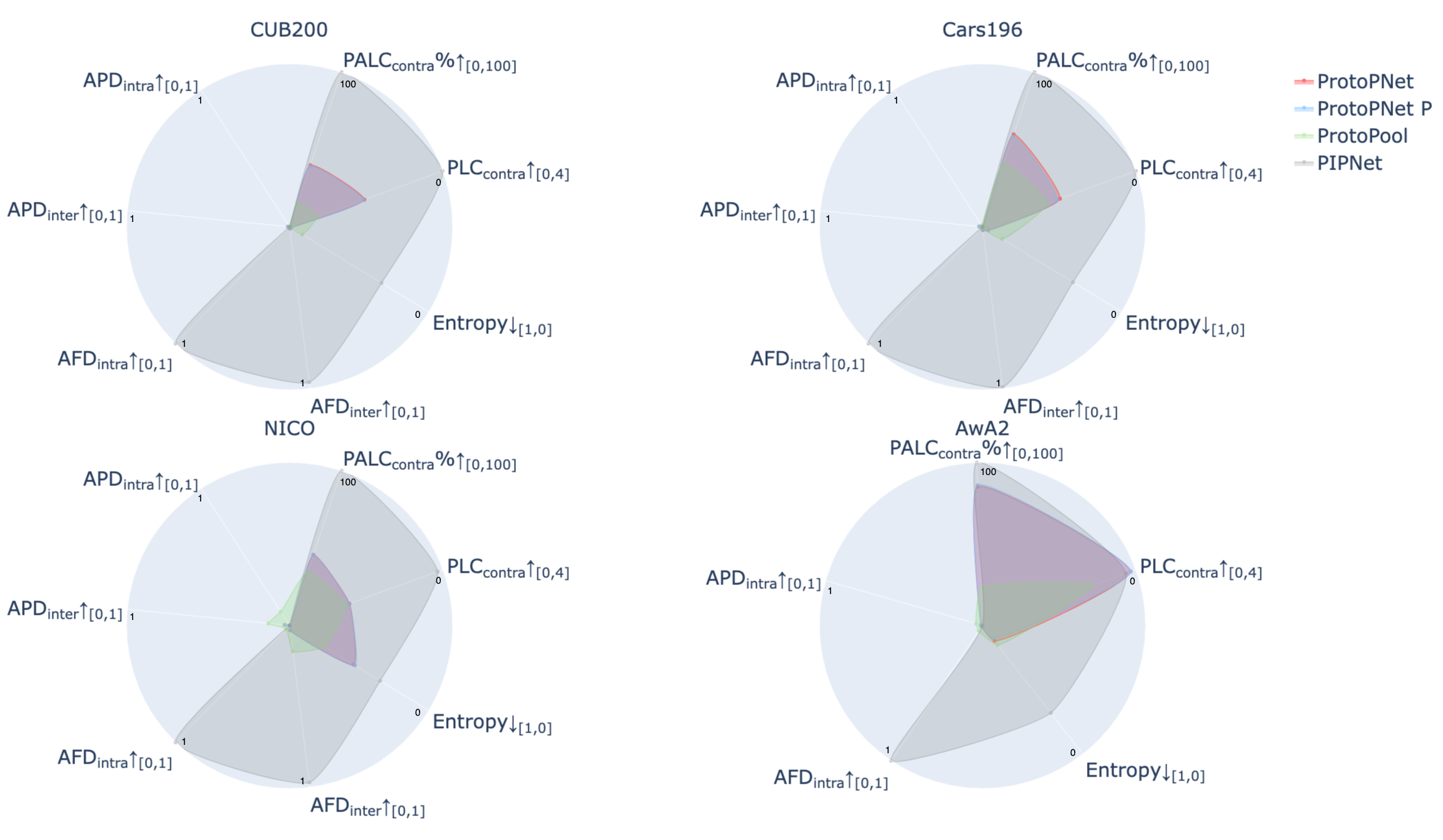}
    \caption{Contrastivity evaluation results. The results are averaged over 4 runs with standard
    deviation. Training and validation subsets were created using 4-fold stratified cross-validation.}
    \label{fig:contrastivity}
\end{figure*}
All contrastivity results in Figure~\ref{fig:contrastivity} were normalized using either fixed metric boundaries or the maximum values observed across models. The metric axes, where lower values indicate better performance, are inverted to consistently increase the coverage area of the better performing models. 

The original and pruned versions of \textit{ProtoPNet} exhibit highly similar behaviours. Both show weak location contrast in the feature maps and low feature and prototype distance scores, indicating that the learned embedding space is not disentangled. Consequently, the extracted features contain similar information rather than distinct, high-level object-part representations. This lack of disentanglement also reduces class separability in the embedding space, suggesting that the backbone network primarily encodes generic shape and colour information rather than interpretable semantic parts. As a result, the embedding space is comparatively smaller and produces ambiguous prototype activations, as reflected by higher entropy values.

\textit{ProtoPool} exhibits characteristics similar to \textit{ProtoPNet}, with an embedding space that appears densely clustered and dominated by generic shape and colour information. This indicates the model’s limiting ability to capture separable high-level features. The attempt to learn shared prototypes seems to exacerbate this issue, compressing the embedding space even further compared to \textit{ProtoPNet}. However, improvements are observed in the \textit{NICO} dataset, where inter-class prototype and feature distances are slightly higher. This suggests that in domains where low-level features (e.g., basic shapes) provide stronger class distinctions, such as differentiating between animal species, \textit{ProtoPool’s} shared prototype approach can be more effective.

\textit{PIPNet} achieves the highest scores in the feature distance measures. Prototype distances are not taken into account due to the lack of explicit prototype vectors. Unlike the other models, \textit{PIPNet} does not treat the backbone outputs as embedding vectors; instead, it interprets each feature dimension as a prototype assignment via softmax. Under this formulation, the feature distance can be understood as the degree of overlap between prototype activations. Including the \textit{PALC}, \textit{PLC}, and \textit{AFD} scores, \textit{PIPNet} shows near-perfect contrast in prototype localization, demonstrating its ability to learn a disentangled feature extraction process. This further indicates that \textit{PIPNet} successfully learns high-level prototypes representing distinct semantic properties. In other words, the extracted features align with the intended goal of representing meaningful and separable object parts. The \textit{Entropy} results reinforce this conclusion, showing that \textit{PIPNet’s} embedding space is more evenly distributed than that of other models. This distribution facilitates discrete prototype-to-class activations, even when prototypes are shared across classes.
\subsection{Continuity}
\begin{figure*}
    \centering
    \includegraphics[width=\linewidth]{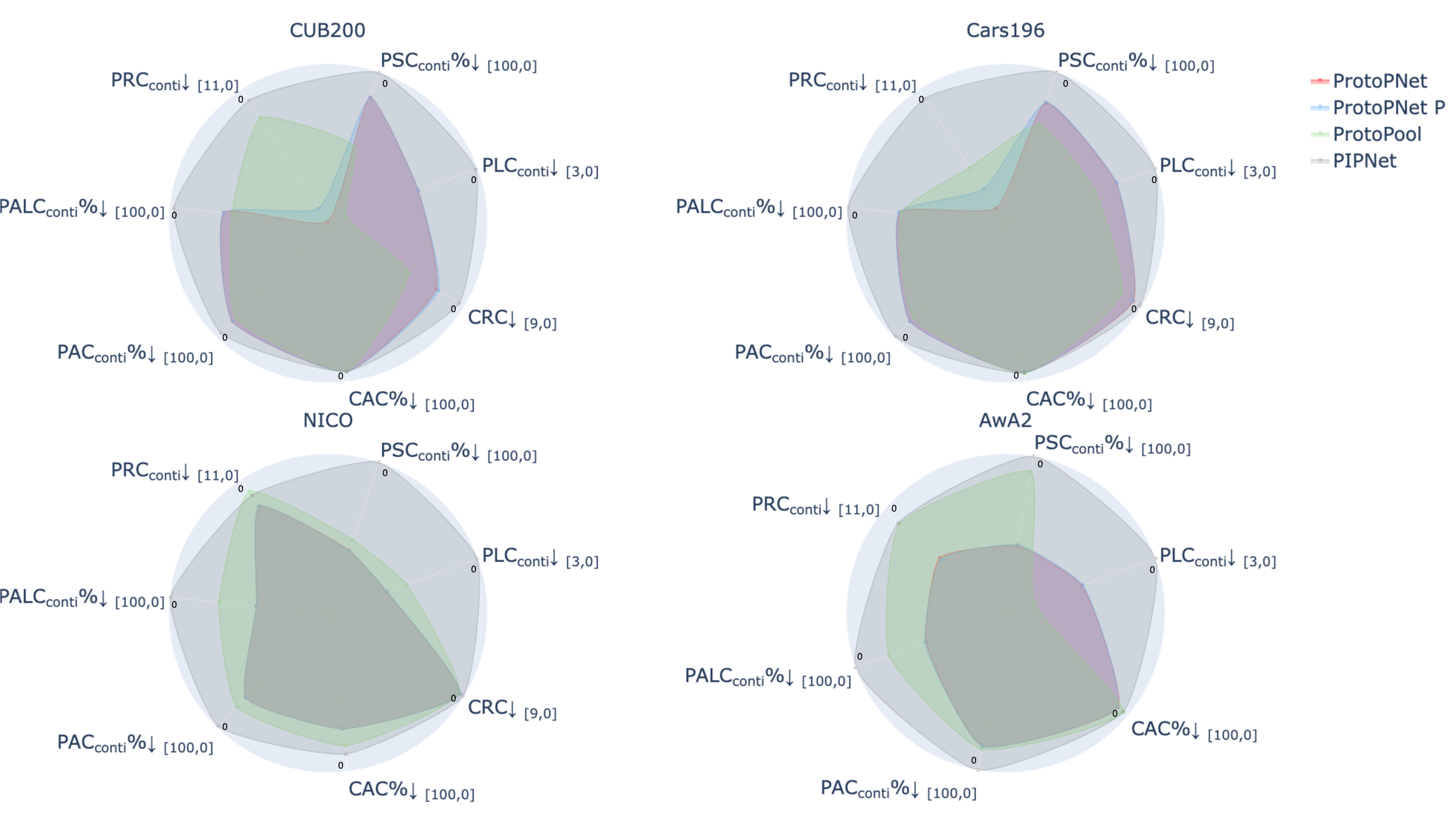}
    \caption{Continuity evaluation results. The results are averaged over 4 runs with standard
    deviation. Training and validation subsets were created using 4-fold stratified cross-validation.}
    \label{fig:continuity}
\end{figure*}
All continuity results in Figure~\ref{fig:continuity} were normalized using either fixed metric boundaries or the maximum values observed across models. Again, we inverted axes of metrics where better performance is indicated by lower scores, so higher performing models cover a larger area in the chart.

The pruned version of \textit{ProtoPNet} achieves slightly improved \textit{PRC} scores on the \textit{CUB200} and \textit{Cars196} datasets compared to the original model. This suggests that pruning successfully removes prototypes more sensitive to low-level perturbations. However, this improvement does not translate into greater robustness at the classification level, as reflected by unchanged \textit{CAC} and \textit{CRC} scores. This observation is consistent with the compactness evaluation, which revealed a complex mixture of relations rather than a straightforward classification process. Interestingly, results on the \textit{NICO} dataset show higher sensitivity to low-level perturbations than on the fine-grained datasets. This indicates that \textit{ProtoPNet} tends to rely on simpler, low-level features when the overall classification task is less challenging. This underscores the absence of a dedicated mechanism that encourages the learning of high-level features.

The design of \textit{ProtoPool} also does not include such a mechanism. However, it appears learning shared prototypes promote the extraction of more robust features under the right circumstances. This is reflected in the superior performance on the \textit{NICO} and \textit{AWA2} dataset compared to \textit{ProtoPNet}. A noteworthy finding is the high \textit{PLC} score relative to other similarity-map metrics, such as \textit{PALC}. This suggests that while the location of maximum activation may vary, overall the broader activation patterns remain relatively stable. Furthermore, \textit{ProtoPool} performs particularly well on the \textit{AWA2} dataset, indicating that the multi-label classification loss contributes to learning more robust shared prototypes. Another key observation is that better \textit{PRC} scores do not necessarily translate into improved \textit{CRC} scores. This again highlights the complex and sensitive nature of the classification process from the trained \textit{ProtoPool} models, which is in agreement with the findings from the compactness evaluation.

\textit{PIPNet} demonstrates the strongest robustness in both prototype location and activation under low-level image augmentations. This result is expected, as the model employs a contrastive learning strategy that explicitly reduces the influence of such perturbations. Notably, this approach also appears to facilitate the extraction of high-level object-part features, as reflected in the contrastive evaluation. Nevertheless, some fluctuations are observed in classification-focused metrics, particularly in the \textit{PRC}. The larger variability in \textit{PRC} compared to the \textit{PSC} suggests that activations among the top-5 prototypes are highly similar. This is likely a consequence of the softmax layer constraining activations between 0 and 1. The sometimes larger \textit{CAC} further reveals that the classification layer amplifies these minor differences. Despite some activation fluctuations, the predicted class remains relatively stable, as indicated by the smaller fluctuations in the \textit{CRC} compared to the \textit{CAC}.

\subsection{Covariate Complexity}
\begin{figure*}[t]
    \centering
    \includegraphics[width=0.44\linewidth]{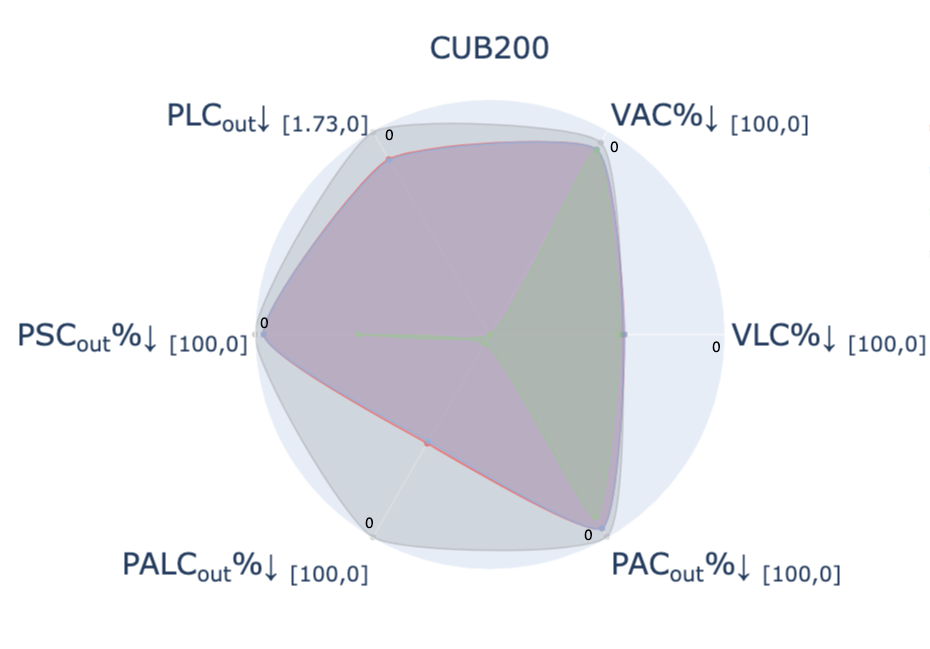}
    \includegraphics[width=0.55\linewidth]{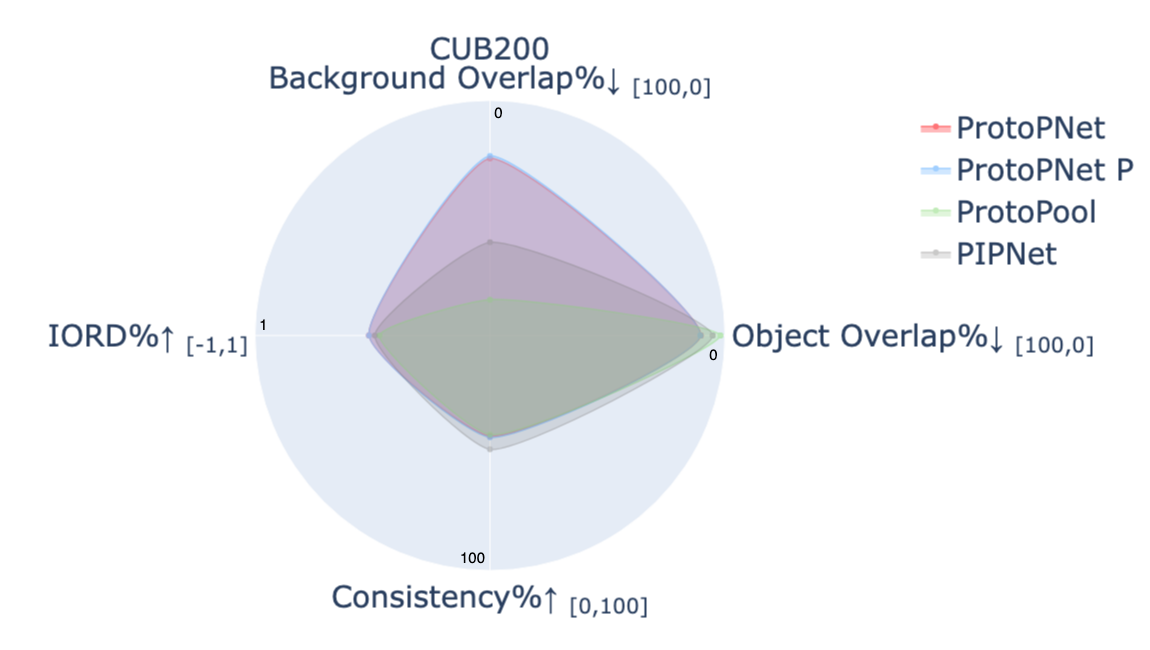}
    \caption{Output-Completeness (left) and Complexity (right) evaluation results on the CUB-200-2011 dataset. The results are averaged over 4 runs with standard
    deviation. Training and validation subsets were created using 4-fold stratified cross-validation.}
    \label{fig:completeness_complexity}
\end{figure*}
All covariate complexity results in Figure~\ref{fig:completeness_complexity} were normalized using either fixed metric boundaries or the maximum values observed across models. The metric axes, where lower values indicate higher performance, are inverted to consistently increase the coverage area of better models.

\textit{ProtoPNet} and its pruned variant produce highly similar results, suggesting that background prototypes were not effectively removed. This finding is consistent with the compactness evaluation, which revealed a complex classification process and indicated that prototypes failed to capture well-defined concepts. Nevertheless, the lower \textit{Background Overlap} compared to \textit{ProtoPool} and \textit{PIPNet} suggests that the class-specific prototypes of the \textit{ProtoPNet} model focus more on object regions. This is intuitive, as backgrounds often serve as shared concepts across classes, and are therefore well suited to be used as shared prototypes.

The \textit{ProtoPool} model exhibits the highest \textit{Background Overlap}, indicating that it is particularly affected by this issue. This observation aligns with the model's overall performance and compactness results on the \textit{CUB200} dataset, which suggest that the learned prototypes lack sufficient class-discriminative information and are of comparatively low quality.

\begin{figure*}[t]
    \centering
    \begin{tabular}{cc|c|c|cc}
        & PIPNet & ProtoPNet & ProtoPNet Pruned & ProtoPool \\
        \midrule
        & (correct pred.) &
        (correct pred.) &
        (correct pred.) &
        (correct pred.) \\
        \rotatebox[origin=l]{90}{\parbox[l]{1.5cm}{1. Proto.}} &
        \includegraphics[width=0.1\linewidth]{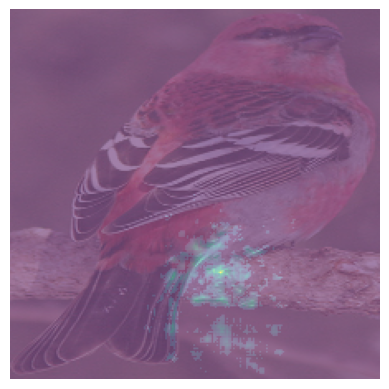} 
        \includegraphics[width=0.1\linewidth]{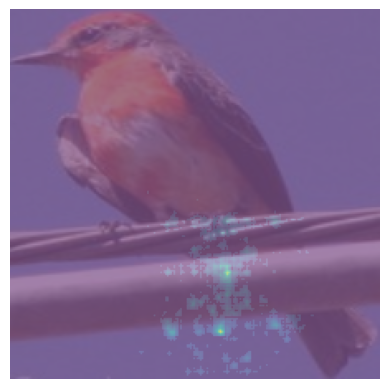} & 
        \includegraphics[width=0.1\linewidth]{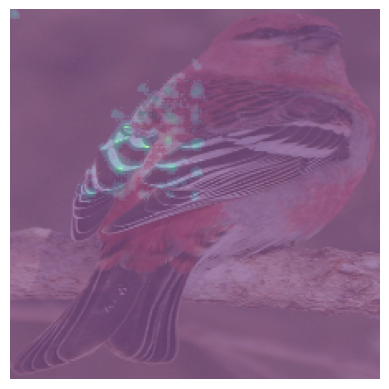}  
        \includegraphics[width=0.1\linewidth]{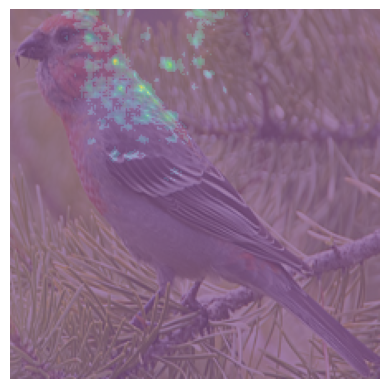} & 
        \includegraphics[width=0.1\linewidth]{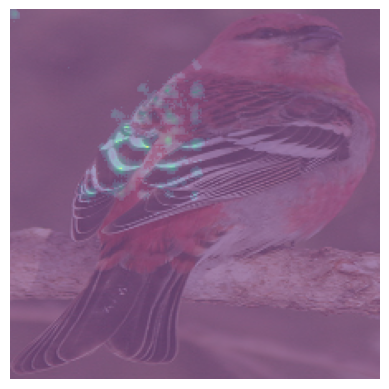}  
        \includegraphics[width=0.1\linewidth]{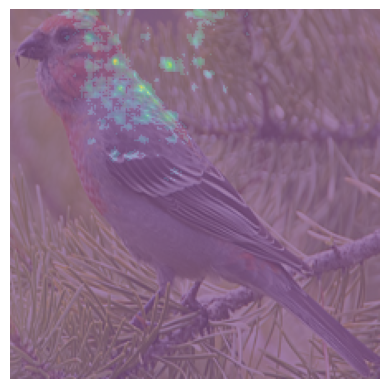} & 
        \includegraphics[width=0.1\linewidth]{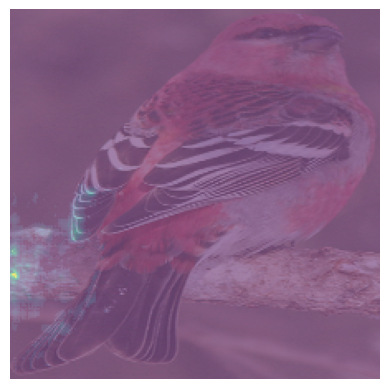} 
        \includegraphics[width=0.1\linewidth]{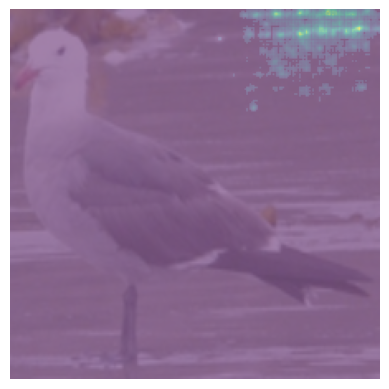} &
        \multirow{2}*{\rotatebox[origin=c]{90}{\parbox[c]{1.4cm}{CUB 200}}} \\
        \rotatebox[origin=l]{90}{\parbox[l]{1.5cm}{2. Proto.}}&
        \includegraphics[width=0.1\linewidth]{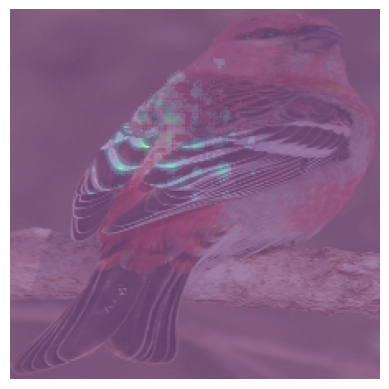} 
        \includegraphics[width=0.1\linewidth]{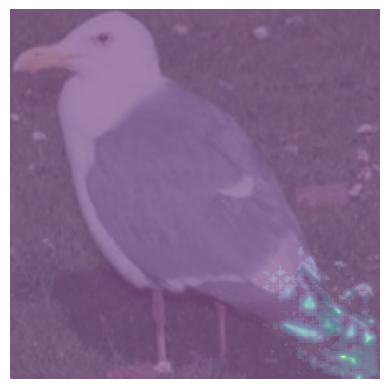} & 
        \includegraphics[width=0.1\linewidth]{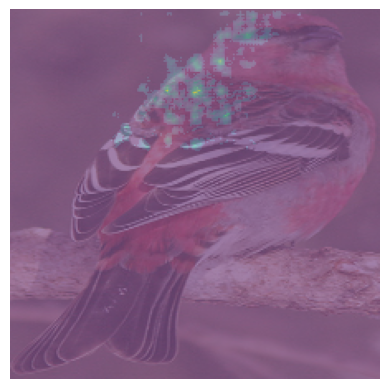}  
        \includegraphics[width=0.1\linewidth]{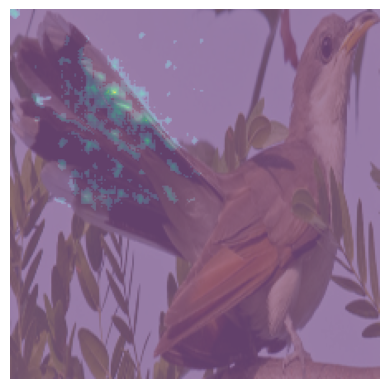} & 
        \includegraphics[width=0.1\linewidth]{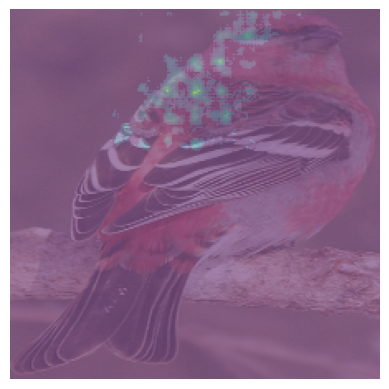}  
        \includegraphics[width=0.1\linewidth]{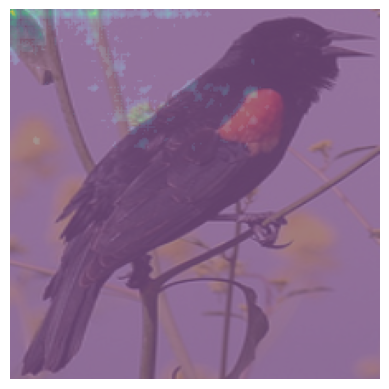} & 
        \includegraphics[width=0.1\linewidth]{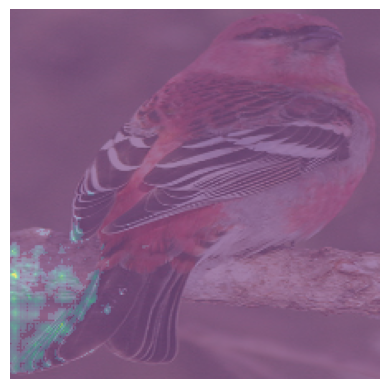}
        \includegraphics[width=0.1\linewidth]{Fig_6_protopnet_cub_train_2.png} \\
        & (correct pred.) &
        (correct pred.) &
        (correct pred.) &
        (correct pred.) \\
        \rotatebox[origin=l]{90}{\parbox[l]{1.5cm}{1. Proto.}} &
        \includegraphics[width=0.1\linewidth]{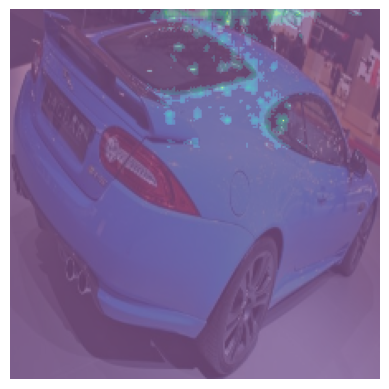} 
        \includegraphics[width=0.1\linewidth]{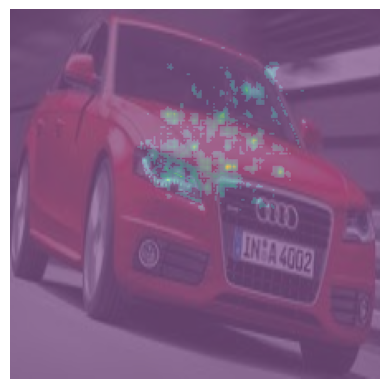} & 
        \includegraphics[width=0.1\linewidth]{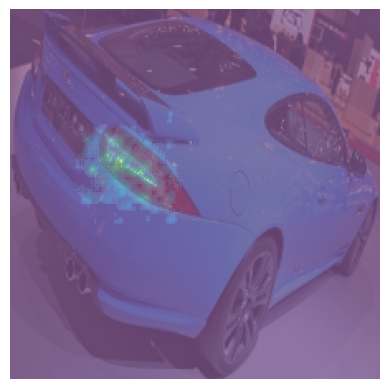}  
        \includegraphics[width=0.1\linewidth]{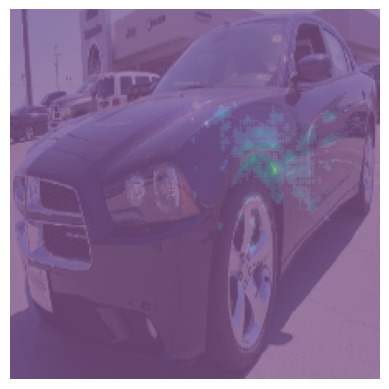} & 
        \includegraphics[width=0.1\linewidth]{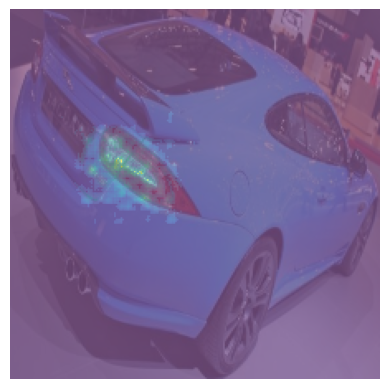}  
        \includegraphics[width=0.1\linewidth]{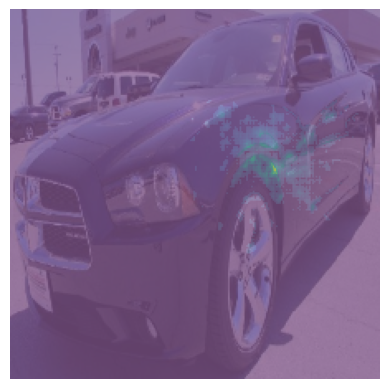} & 
        \includegraphics[width=0.1\linewidth]{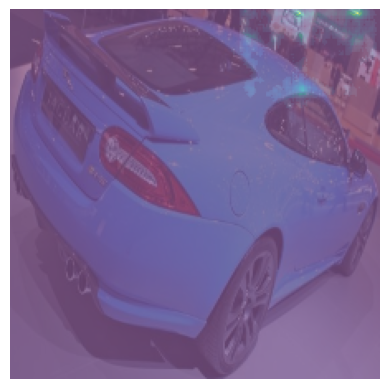} 
        \includegraphics[width=0.1\linewidth]{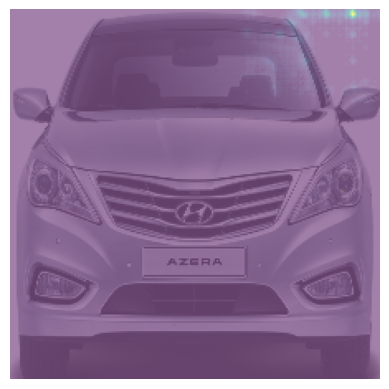} &
        \multirow{2}*{\rotatebox[origin=c]{90}{\parbox[c]{0.7cm}{Cars}}} \\
        \rotatebox[origin=l]{90}{\parbox[l]{1.5cm}{2. Proto.}}&
        \includegraphics[width=0.1\linewidth]{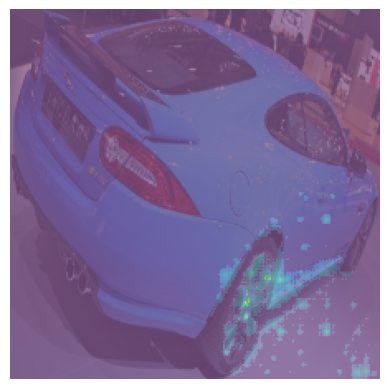} 
        \includegraphics[width=0.1\linewidth]{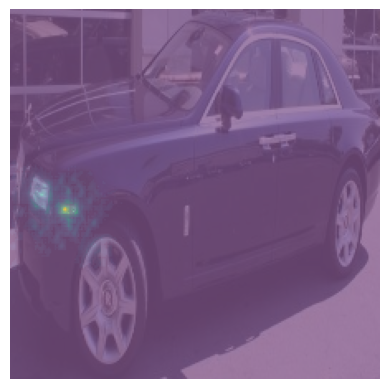} & 
        \includegraphics[width=0.1\linewidth]{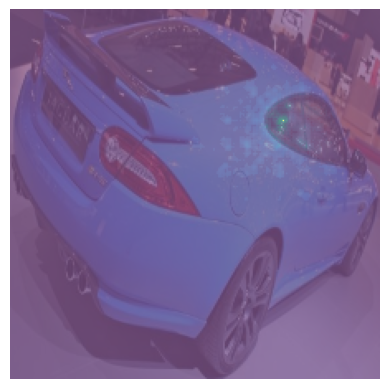}  
        \includegraphics[width=0.1\linewidth]{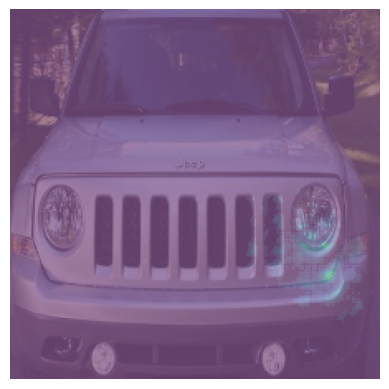} & 
        \includegraphics[width=0.1\linewidth]{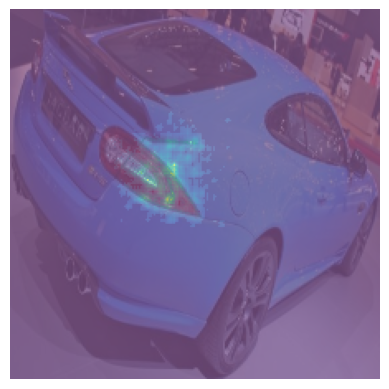}  
        \includegraphics[width=0.1\linewidth]{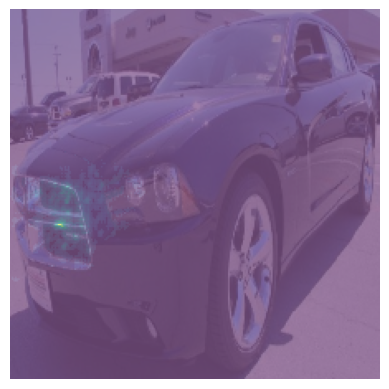} & 
        \includegraphics[width=0.1\linewidth]{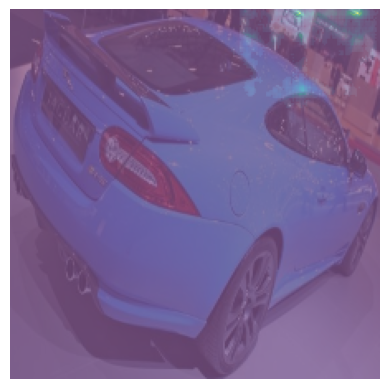}
        \includegraphics[width=0.1\linewidth]{Fig_6_protopnet_cars_train_2.png} \\
        & \includegraphics[width=0.13\linewidth]{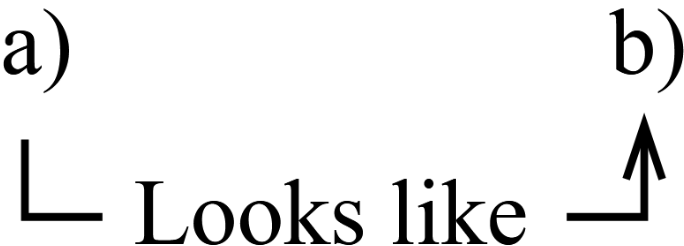}&
        \includegraphics[width=0.13\linewidth]{Fig_6_7_looks_like.png}&
        \includegraphics[width=0.13\linewidth]{Fig_6_7_looks_like.png}&
        \includegraphics[width=0.13\linewidth]{Fig_6_7_looks_like.png}\\
    \end{tabular}
    \caption{Example prototype saliency maps using the \textit{PRP} visualization method on the \textit{CUB200} (top) and \textit{Cars196} (bottom) dataset. The visualizations show the top-2 prototypes with the highest similarity score. Images in column a) illustrate a sample from the test set. The column b) visualizes the prototype on the nearest training image, respectively. Each saliency map visualizes only the 95th percentile.}
    \label{fig:cub_cars_examples}
\end{figure*}

Interestingly, the \textit{IORD} scores are nearly identical across all models, indicating that, on average, model focus is balanced between object and background regions despite differing levels of \textit{Background Overlap}. This suggests that prototype activations are relatively uniform, probably due to the 95th percentile crop, leading to similar average scores for both object- and background-overlapping activations. This can also be observed in the example visualization in Figure~\ref{fig:cub_cars_examples} and \ref{fig:nico_awa2_examples}.

Object-part \textit{Consistency} is generally low across models, with \textit{PIPNet} performing slightly better. This indicates that prototypes in all models tend to capture multiple object parts rather than maintaining a one-to-one correspondence between prototypes and individual semantic parts, which diverges from the intended interpretability objective of prototype-based learning. On average, prototypes focus on approximately one-tenth of the object, as measured by the \textit{Object Overlap}. This indicates that while a prototype may represent a coherent concept within a single sample image, it typically corresponds to multiple concepts across the test set.

\subsection{Output-Completeness}
All output-completeness results in Figure~\ref{fig:completeness_complexity} were normalized using either fixed metric boundaries or the maximum values observed across models. As previously mentioned, metric axes are inverted when lower values indicate better performance.

The evaluation of the \textit{PRP} method for visualizing relevant pixels yields mixed results across all models. The visualizations are most consistent with the feature extraction process of \textit{PIPNet}, which aligns best with the intended interpretation of high-level object-part extraction, as shown in previous evaluations. This suggests a clear correlation between the accuracy of \textit{PRP} visualizations and the degree of disentanglement in the extracted features.
The second-best visualizations were done on \textit{ProtoPNet}, although with greater variability in activation location metrics such as \textit{PLC} and \textit{PALC}. This indicates that \textit{ProtoPNet} prototypes are influenced by broader image regions compared to \textit{PIPNet}, which the \textit{PRP} method does not fully capture. Examples of \textit{PRP} visualizations are illustrated in Figure~\ref{fig:cub_cars_examples} and \ref{fig:nico_awa2_examples},which show similar sizes of the relevance areas. A similar trend is observed for \textit{ProtoPool}, where visualizations are less reliable due to pronounced prototype location instability (as seen in the continuity evaluation) and a similar weak inter-prototype contrast, resulting in a more entangled feature space.
Overall, these findings suggest that the effectiveness of backpropagation-based visualization methods is strongly dependent on the disentanglement of the learned embedding space.

\section{Discussion}
\begin{figure*}[t]
    \centering
    \begin{tabular}{cc|c|c|cc}
        & PIPNet & ProtoPNet & ProtoPNet Pruned & ProtoPool \\
        \midrule
        & (TP 11/17 TN 32/32) &
        (TP 17/17 TN 32/32) &
        (TP 17/17 TN 32/32) &
        (TP 17/17 TN 30/32) \\
        \rotatebox[origin=l]{90}{\parbox[l]{1.5cm}{1. Proto.}} &
        \includegraphics[width=0.1\linewidth]{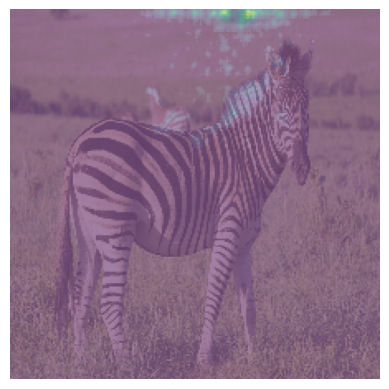} 
        \includegraphics[width=0.1\linewidth]{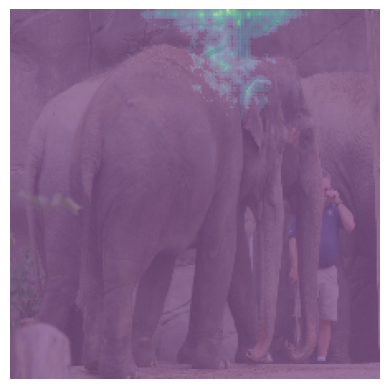} & 
        \includegraphics[width=0.1\linewidth]{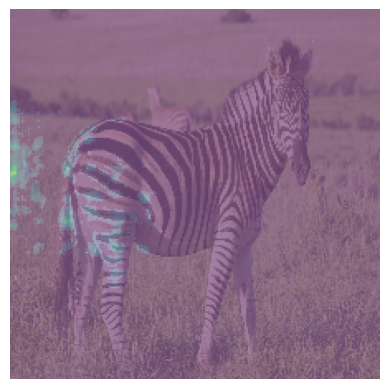}  
        \includegraphics[width=0.1\linewidth]{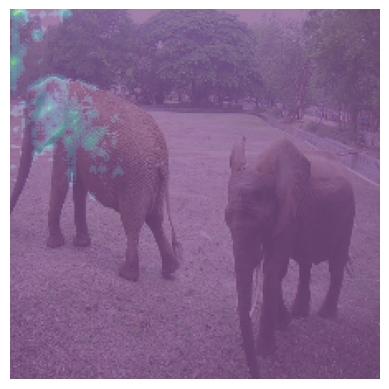} & 
        \includegraphics[width=0.1\linewidth]{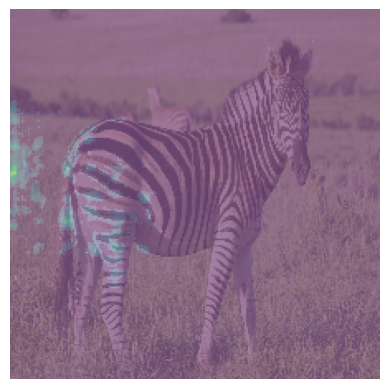}  
        \includegraphics[width=0.1\linewidth]{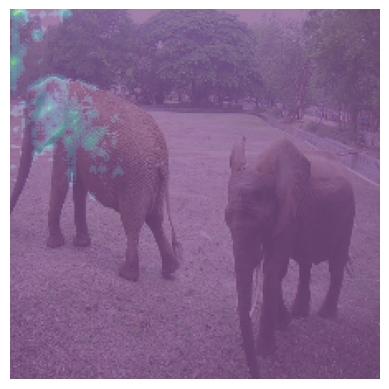} & 
        \includegraphics[width=0.1\linewidth]{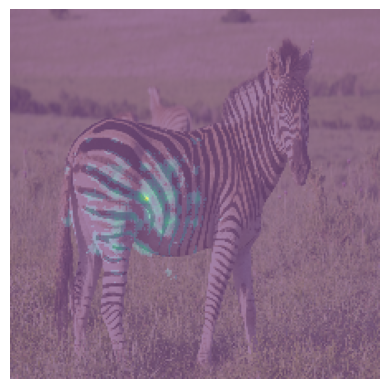} 
        \includegraphics[width=0.1\linewidth]{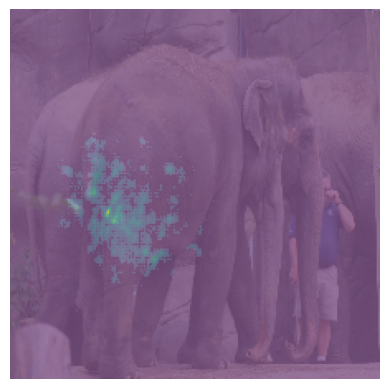} &
        \multirow{2}*{\rotatebox[origin=c]{90}{\parbox[c]{0.8cm}{AwA2}}} \\
        \rotatebox[origin=l]{90}{\parbox[l]{1.5cm}{2. Proto.}}&
        \includegraphics[width=0.1\linewidth]{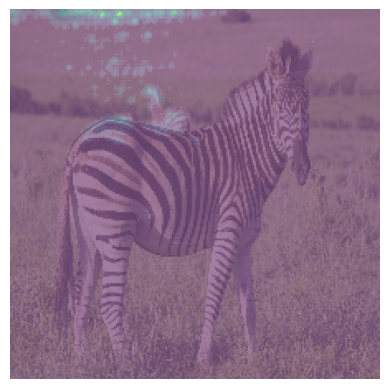} 
        \includegraphics[width=0.1\linewidth]{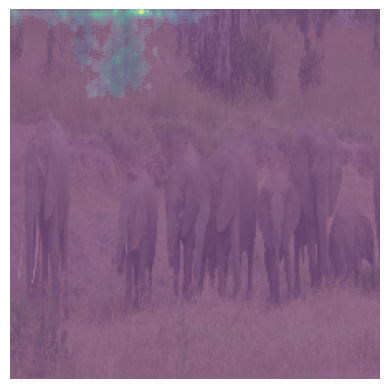} & 
        \includegraphics[width=0.1\linewidth]{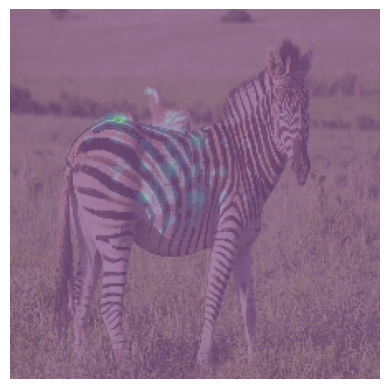}  
        \includegraphics[width=0.1\linewidth]{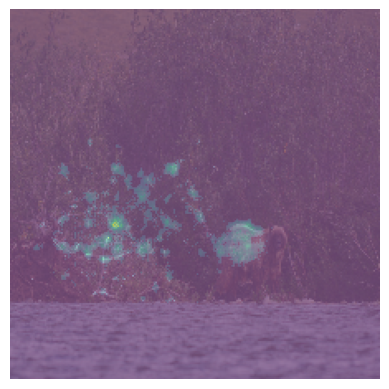} & 
        \includegraphics[width=0.1\linewidth]{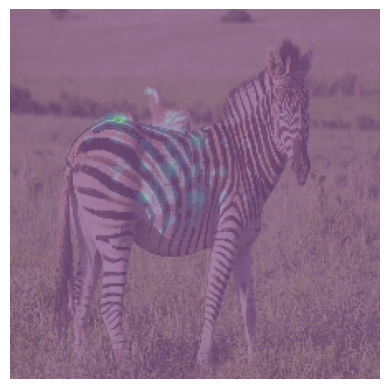}  
        \includegraphics[width=0.1\linewidth]{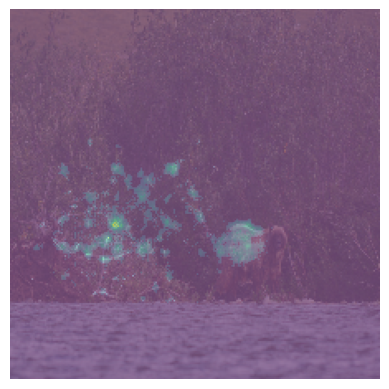} & 
        \includegraphics[width=0.1\linewidth]{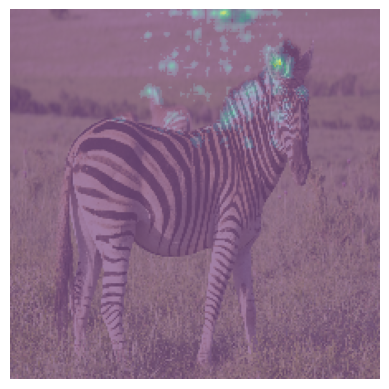}
        \includegraphics[width=0.1\linewidth]{Fig_7_protopnet_awa2_train_2.png} \\
        & (correct pred.) &
        (correct pred.) &
        (correct pred.) &
        (correct pred.) \\
        \rotatebox[origin=l]{90}{\parbox[l]{1.5cm}{1. Proto.}} &
        \includegraphics[width=0.1\linewidth]{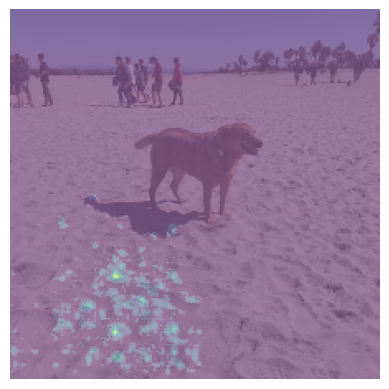} 
        \includegraphics[width=0.1\linewidth]{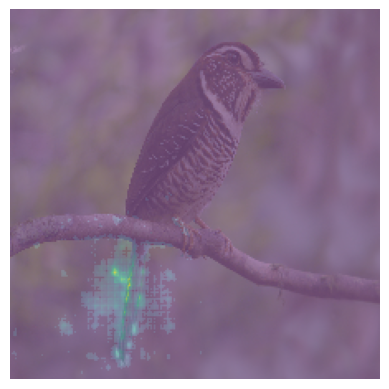} & 
        \includegraphics[width=0.1\linewidth]{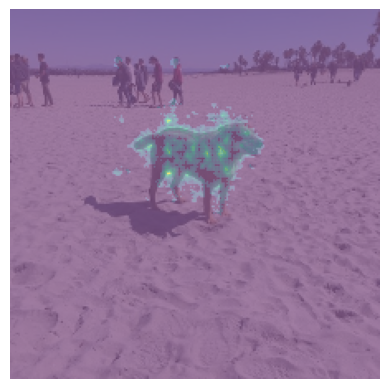}  
        \includegraphics[width=0.1\linewidth]{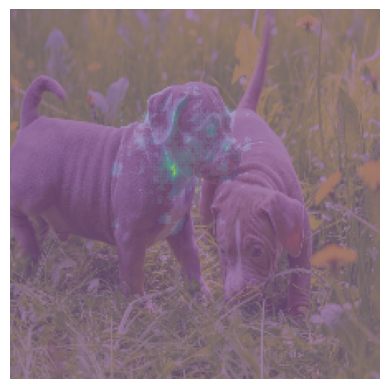} & 
        \includegraphics[width=0.1\linewidth]{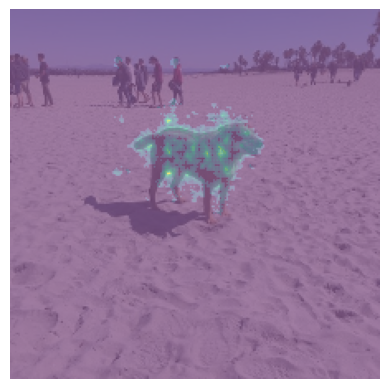}  
        \includegraphics[width=0.1\linewidth]{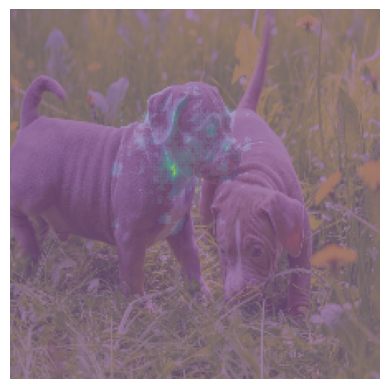} & 
        \includegraphics[width=0.1\linewidth]{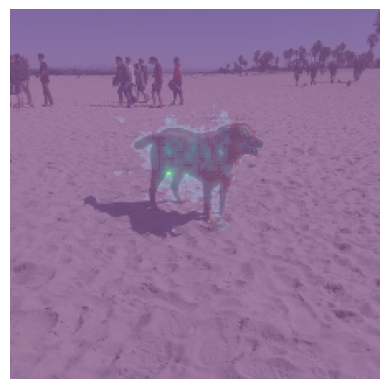} 
        \includegraphics[width=0.1\linewidth]{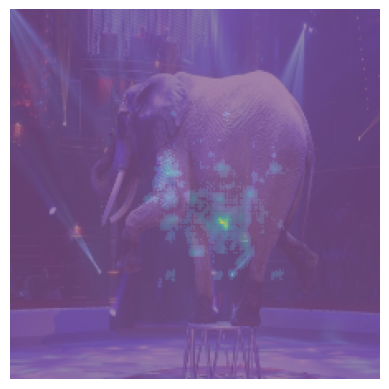} &
        \multirow{2}*{\rotatebox[origin=c]{90}{\parbox[c]{0.7cm}{Nico}}} \\
        \rotatebox[origin=l]{90}{\parbox[l]{1.5cm}{2. Proto.}}&
        \includegraphics[width=0.1\linewidth]{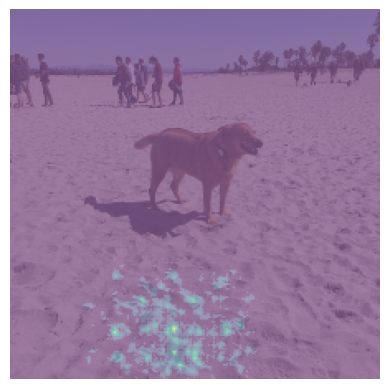} 
        \includegraphics[width=0.1\linewidth]{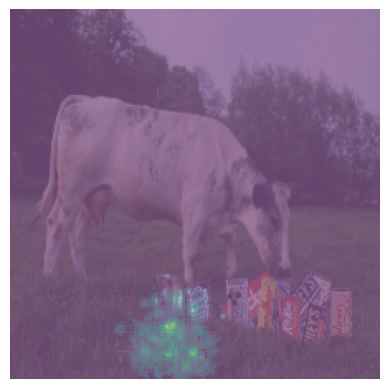} & 
        \includegraphics[width=0.1\linewidth]{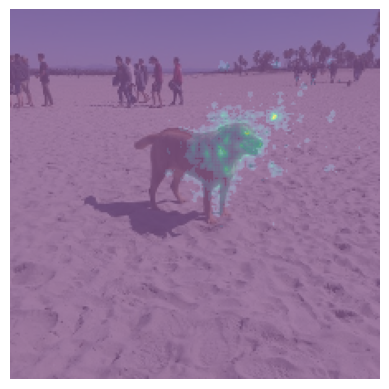}  
        \includegraphics[width=0.1\linewidth]{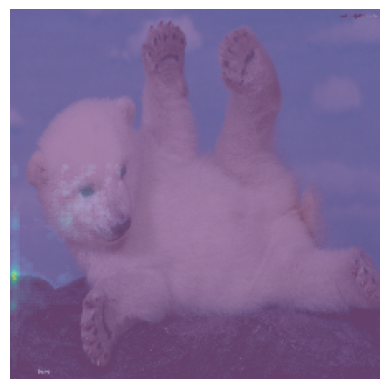} & 
        \includegraphics[width=0.1\linewidth]{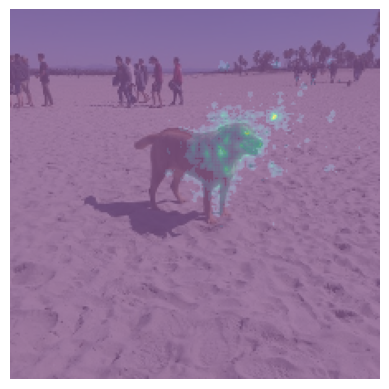}  
        \includegraphics[width=0.1\linewidth]{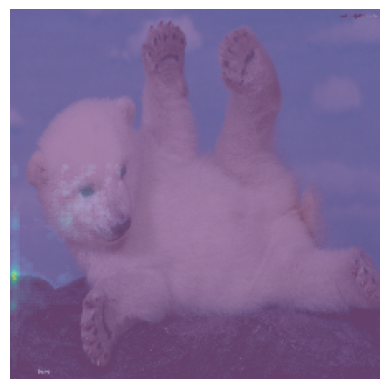} & 
        \includegraphics[width=0.1\linewidth]{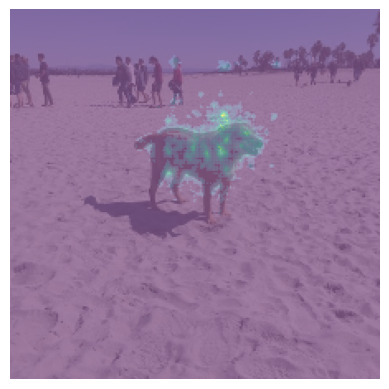}
        \includegraphics[width=0.1\linewidth]{Fig_7_protopnet_nico_train_2.png} \\
        & \includegraphics[width=0.13\linewidth]{Fig_6_7_looks_like.png}&
        \includegraphics[width=0.13\linewidth]{Fig_6_7_looks_like.png}&
        \includegraphics[width=0.13\linewidth]{Fig_6_7_looks_like.png}&
        \includegraphics[width=0.13\linewidth]{Fig_6_7_looks_like.png}\\
    \end{tabular}
    \caption{Example prototype saliency maps using the \textit{PRP} visualization method on the \textit{AWA2} (top) and \textit{NICO} (bottom) dataset. The visualizations show the top-2 prototypes with the highest similarity score. Images in column a) illustrate a sample from the test set. The column b) visualizes the prototype on the nearest training image, respectively. Each saliency map visualizes only the 95th percentile.}
    \label{fig:nico_awa2_examples}
\end{figure*}
\textit{ProtoPNet} primarily learned complex prototypes that captured low-level image characteristics rather than the intended semantic object parts, leading to an equally complex classification process. Pruning removed a small number of prototypes and slightly improved interpretability, but consistently deleting background prototypes could not be achieved.

Example visualizations on the fine-grained datasets (Figure~\ref{fig:cub_cars_examples}) illustrate the challenges in interpreting the concepts learned by individual prototypes. While the learned concept can often be guessed when examining a single image, comparisons between the nearest feature visualizations from the training set and the sample image reveal that a prototype often encodes multiple concepts simultaneously. This issue is particularly pronounced in the other datasets (Figure~\ref{fig:nico_awa2_examples}). For example, in the \textit{AWA2} dataset, interpretation is especially vague, we can hardly guess prototype concepts such as “stripes” in zebras, while in the \textit{NICO} dataset, sample features often span much larger regions than their nearest training features.

On \textit{CUB200} and \textit{Cars196}, our accuracies are 5–10\% below the original reports, most likely due to halving prototype channels (256→128) for more stability under our training setup. In addition, only the classification loss was class-weighted to counter imbalance; cluster/separation losses were not. Pruning produced only marginal gains in interpretability, falling short of the clearer pruning effects reported by Chen et al. \cite{chen2019thislookslikethat}. Compared to Wang et al. \cite{wang2023learning}, our PRP-based visualizations yield smaller, sparser activations than the original upsampling method, explaining their higher \textit{Object Overlap} measures.

\textit{ProtoPool} exhibited prototypes and classification processes of varying complexity, with clear differences between the two fine-grained datasets. Its performance appears to be strongly influenced by the complexity of shared object parts between classes. For example, in \textit{Cars196}, shared car parts such as tires can be represented with low-level features like colour and shape, whereas in \textit{CUB200}, the shared relations among bird species are more complex. Overall, \textit{ProtoPool} tends to learn low-level image characteristics but emphasizes simpler, more interpretable classification processes than \textit{ProtoPNet}, particularly in less complex classification tasks.
We saw surprisingly good performance on the challenging multi-label dataset, suggesting that the shared prototype approach aligns well with such applications.

Figure~\ref{fig:cub_cars_examples} illustrates that prototypes in both fine-grained datasets often focus heavily on the background. Even in the better-performing \textit{Cars196} dataset, this issue is clearly visible. In contrast, visualizations on \textit{NICO} and \textit{AWA2} (Figure~\ref{fig:nico_awa2_examples}) demonstrate improved object focus, comparable to \textit{ProtoPNet}, which is consistent with \textit{ProtoPool’s} stronger results on these datasets.

Similar to \textit{ProtoPNet}, our obtained \textit{ProtoPool} scores on \textit{CUB200} and \textit{Cars196} are 5–10\% below the original, that we explain by the same channel reduction and class-imbalance handling.

\textit{PIPNet} achieved the best results in both prototype quality and classification simplicity, making it the most interpretable model. The combination of the softmax layer with the tanh loss proved highly effective in learning a minimal yet sufficient set of prototypes to achieve high accuracy. Furthermore, design choices in the classification layer, especially the hard positive constraint and the adapted output computation, successfully reduced the \textit{Local Size}, yielding a more interpretable classification process compared to the other models.

However, visual explanations (Figures~\ref{fig:cub_cars_examples} and \ref{fig:nico_awa2_examples}) reveal issues similar to those observed in \textit{ProtoPNet} and \textit{ProtoPool}. While prototype concepts can often be guessed in single-image visualizations, the nearest feature comparisons again reveal mismatches. Interestingly, \textit{PIPNet} focuses more on background regions in the \textit{NICO} and \textit{AWA2} datasets, showing the opposite trend of \textit{ProtoPool}, despite both models employing shared prototypes.
Relative to Nauta et al. \cite{nauta2023pip}, our \textit{PIPNet} models underperform by 8\%, which we primarily attribute to the lower feature map resolution used here ($7{\times}7$ vs. $28{\times}28$) to ensure a uniform comparison across methods. Despite this gap, our compactness results mirror the original findings: comparable reductions in global size and high sparsity in the classification layer. \textit{PRP} visualizations align best with \textit{PIPNet}, though object-part consistency drops notably compared to the purity metric from the original study that focuses only on the top-10 nearest training images. On \textit{AWA2}, \textit{PIPNet} attains the highest accuracy. However, performance is likely suppressed by highly imbalanced labels and a multi-label margin loss that cannot incorporate class weights.

Overall, our evaluation focused on diverse model aspects relevant to interpretability, with mixed results across models and reappearing patterns in different evaluation techniques from individual architectures.
We see that models generally struggle to learn semantically distinct prototypes and are more focused on lower level image characteristics and shapes. 
This often results in a cramped latent space and similar activation locations, facilitating a complex relation between prototypes and classes.
It is evident that learning semantic prototypes is a complex task, that current architectures struggle to consistently achieve. Other approaches that promote some of the evaluated properties in a more direct matter like  
contrastive learning approaches seem to be able to mitigate this problem well, as seen by the \textit{PIPNet} model.

\section{Conclusion}
We examined the evaluation of prototype-based neural networks, drawing on the foundational research by Nauta and Seifert~\cite{nauta2023co12} and Nauta et al. \cite{nauta2023anecdotal}. The Co-12 properties identified by Nauta et al. \cite{nauta2023anecdotal} provide a robust framework for assessing the quality of explanations in XAI methods. We complement these using several new metrics and provide a comprehensive evaluation using four datasets. Our python library integrates the existing evaluation techniques in the Co-12 framework and offers a practical toolkit for validating, benchmarking, and comparing part-based prototype networks.

For future work, we aim to apply this library for developing and refining new prototype models, in particular for (fine-)tuning and evaluation. In addition, we aim to extend the experimentation on further datasets.

\section*{Statements and Declarations}
All authors certify that they have no affiliations with or involvement in any organization or entity with any financial interest or non-financial interest in the subject matter or materials discussed in this manuscript.

\section*{Acknowledgments}
This work has been partially supported by the funded project FRED, German Federal Ministry for Economic Affairs and Climate Action (BMWK), FKZ: 01MD22003E, as well as by the German Research Foundation (DFG), grant MODUS-II (316679917, AT 88/4-2).

\section*{Declarations}
\textbf{Competing Interests}\\
The authors declare that they have no competing interests related to the subject matter discussed in this manuscript.\\

\noindent
\textbf{Data Availability}\\
The datasets generated and analyzed during the current study are publicly available. Further information, including access links, can be found in the repository at https://github.com/uos-sis/quanproto.
\bibliography{sn-bibliography}
\end{document}